\documentclass[journal]{IEEEtran}
\usepackage{graphicx}
\usepackage{cite}
\usepackage{epstopdf}
\usepackage{amsfonts}
\usepackage{amsmath}
\usepackage{algorithm}
\usepackage{algorithmic}
\usepackage{subfig}
\usepackage{array}
\usepackage{multicol}
\usepackage{multirow}
\usepackage{enumerate}
\usepackage{color}

\begin{document}

\title{Semantics-aware Adaptive Knowledge Distillation for Sensor-to-Vision Action Recognition}

\author{Yang~Liu,~\IEEEmembership{Member,~IEEE}, Keze~Wang, Guanbin~Li,~\IEEEmembership{Member,~IEEE}, Liang~Lin,~\IEEEmembership{Senior~Member,~IEEE}
\thanks{This work is supported in part by the National Natural Science Foundation of China under Grant No.62002395, in part by the National Natural Science Foundation of Guangdong Province (China) under Grant No. 2021A15150123, and in part by the China Postdoctoral Science Foundation funded project under Grant No.2020M672966. (\emph{Corresponding author: Liang Lin.})}
\thanks{Yang Liu, Guanbin Li and Liang Lin are with the School of Computer Science and Engineering, Sun Yat-Sen University, Guangzhou 510006, China. (e-mail: liuy856@mail.sysu.edu.cn, liguanbin@mail.sysu.edu.cn, linliang@ieee.org).

Keze Wang is with DMAI Co., Ltd, Guangzhou 511400, China. (e-mail: kezewang@gmail.com).}}

\markboth{ IEEE TRANSACTIONS ON IMAGE PROCESSING}%
{Shell \MakeLowercase{\textit{et al.}}: Bare Demo of IEEEtran.cls for IEEE Journals}

\maketitle

\begin{abstract}
Existing vision-based action recognition is susceptible to occlusion and appearance variations, while wearable sensors can alleviate these challenges by capturing human motion with one-dimensional time-series signals (e.g. acceleration, gyroscope, and orientation). For the same action, the knowledge learned from vision sensors (videos or images) and wearable sensors, may be related and complementary. However, there exists a significantly large modality difference between action data captured by wearable-sensor and vision-sensor in data dimension, data distribution, and inherent information content. In this paper, we propose a novel framework, named Semantics-aware Adaptive Knowledge Distillation Networks (SAKDN), to enhance action recognition in vision-sensor modality (videos) by adaptively transferring and distilling the knowledge from multiple wearable sensors. The SAKDN uses multiple wearable-sensors as teacher modalities and uses RGB videos as student modalities. To preserve the local temporal relationship and facilitate employing visual deep learning models, we transform one-dimensional time-series signals of wearable sensors to two-dimensional images by designing a gramian angular field based virtual image generation model. Then, we introduce a novel Similarity-Preserving Adaptive Multi-modal Fusion Module (SPAMFM) to adaptively fuse intermediate representation knowledge from different teacher networks. Finally, to fully exploit and transfer the knowledge of multiple well-trained teacher networks to the student network, we propose a novel Graph-guided Semantically Discriminative Mapping (GSDM) module, which utilizes graph-guided ablation analysis to produce a good visual explanation to highlight the important regions across modalities and concurrently preserve the interrelations of original data. Experimental results on Berkeley-MHAD, UTD-MHAD, and MMAct datasets well demonstrate the effectiveness of our proposed SAKDN for adaptive knowledge transfer from wearable-sensors modalities to vision-sensors modalities. The code is publicly available at https://github.com/YangLiu9208/SAKDN.
\end{abstract}


\IEEEpeerreviewmaketitle

\section{Introduction}

\IEEEPARstart{H}{uman} action recognition has attracted increasing attention due to its wide applications such as health-care services, smart homes, intelligent surveillance, and human-machine interaction, etc. With the development of deep learning, vision-sensors (images, videos) based methods dominate the community of action recognition and a large amount of effective models have been proposed and applied to real-world scenarios \cite{wang2018temporal,zhou2018temporal,wang2018non, liu2021temporal}. However, the performance of vision-based methods is easily affected by camera position, camera view-point, background clutter, occlusion, and appearance variation \cite{kong2019mmact}. Furthermore, vision-based methods usually require expensive hardware resources to run computationally complex computer vision algorithms \cite{ehatisham2019robust}. In some privacy-sensitive areas such as bank and government, the difficulty of acquiring images and videos makes this method infeasible. However, these limitations can be addressed by low-cost and computationally efficient wearable-sensors. The wearable-sensors equipped by smartwatches or smartphones can capture human actions by three-axis time-series acceleration, gyroscope, and orientation signals, which are suitable for privacy protecting and robust to variant illuminations and camera viewpoints \cite{kong2019mmact}. With the popularity and increasing demand of intelligent cities and smart health-care, human action recognition based on wearable-sensors has become a key research area in human activity understanding. Although some wearable-sensors based action recognition methods \cite{jiang2015human,wannenburg2016physical,wang2019deep,wang2019attention} have been proposed and achieved promising results, most of them just consider time-series data of wearable-sensors without considering the complementary relationship and domain divergence between vision-sensor and wearable-sensor action data. Therefore, it is considerable to leverage the knowledge from both vision-sensor and wearable-sensor modalities to improve the performance of action recognition in such a multi-modal manner.

\begin{figure}[!t]
\centering
\includegraphics[scale=0.14]{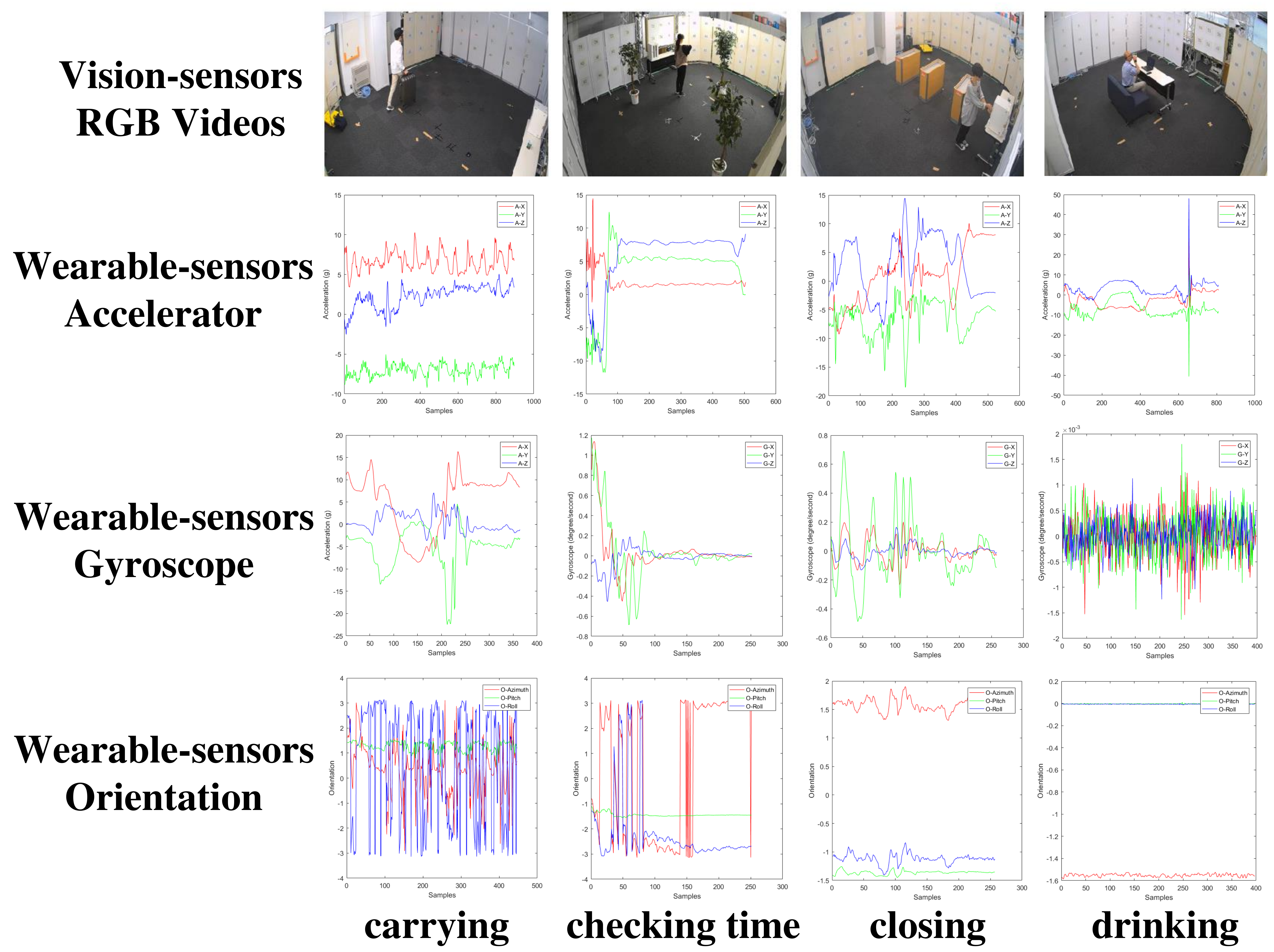}
\caption{Comparison of  vision and wearable sensor action data. }
\vspace{-10pt}
\label{Fig1}
\end{figure}

\begin{figure*}[!t]
\centering
\includegraphics[scale=0.255]{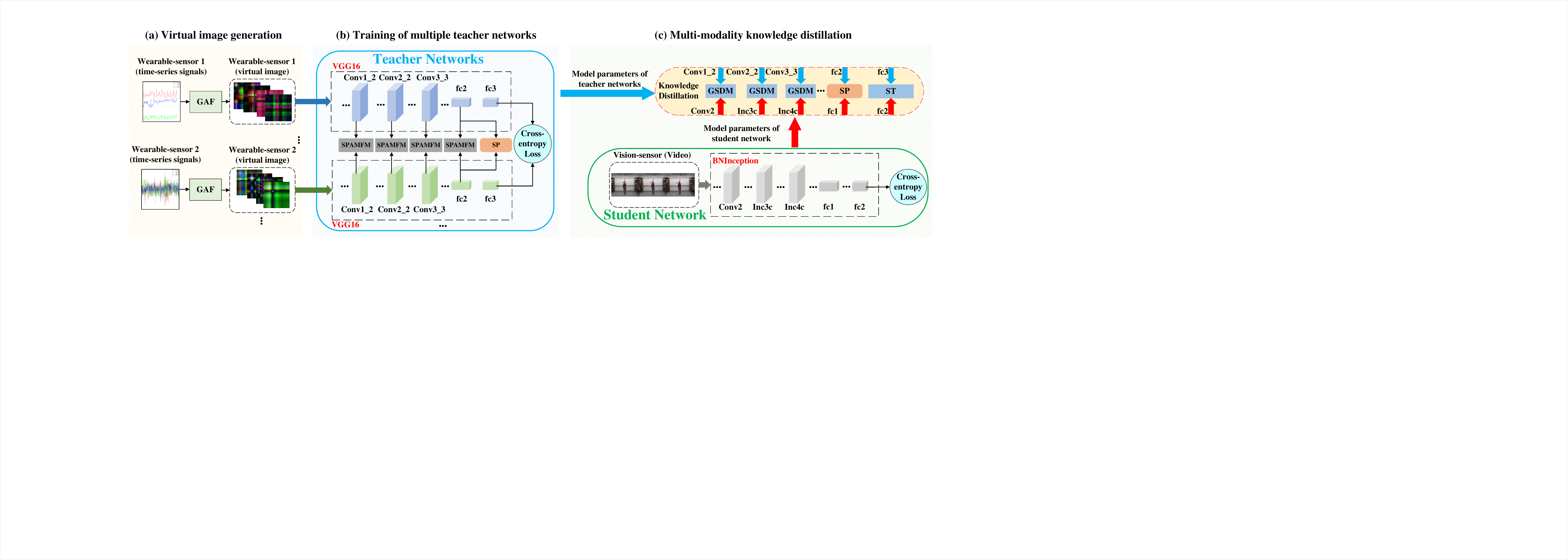}
\caption{Framework of our proposed method SAKDN. (a) Virtual image generation: building a Gramian Angular Field (GAF) based virtual image generation model to transform the one-dimensional time-series signals into two-dimensional virtual image representations. (b) Training of multiple teacher networks: constructing a Similarity-Preserving Adaptive Multi-modal Fusion Module (SPAMFM) that utilizes the intra-modality similarity, semantic embedding, and multiple relational knowledge to fully utilize the complementary information among different teacher networks. (c) Multi-modality knowledge distillation: training the student network using GSDM loss, soft-target loss along with semantic preserving loss, and cross-entropy loss.}
\vspace{-10pt}
\label{Fig2}
\end{figure*}

However, there exists a significantly large modality difference between vision-sensor and wearable-sensor action data, which can be observed from  Fig. \ref{Fig1}. Obviously, the vision-sensor action data are two-dimensional images or three-dimensional videos which contain abundant color or texture information. In contrast, wearable-sensor action data are one-dimensional time-series signals without containing color and texture information. Specially, traditional action recognition methods are usually in a unimodal manner (either in vision-sensor modality or wearable-sensor modality), which is infeasible in real-world scenarios because the dynamic environment makes the model hard to adapt to the modality difference. Previous works \cite{ahmad2019human,dawar2018data,wei2019fusion,garcia2019learning} have verified the existence of complementary information between action data of vision-sensor and wearable-sensor. For instance, vision-based sensors could provide global motion features while wearable-sensors give 3D information about local body movement. Hence, by utilizing the complementary information from these two modalities, the generalization ability and the performance of action recognition can be improved. However, due to the huge modality gap between vision-sensor and wearable-senor action data, the following two key challenges should be addressed: 1) there are multiple modalities of wearable-sensor action data, the data of each modality is one-dimensional time-series signal without containing local temporal relationship, color, and texture information. This makes existing models difficult to interpret and fuse the content of multi-modal wearable-sensor action data. Therefore, a specific and effective multi-modal representation learning method is required to increase the representative power of wearable-sensor data and concurrently fuse different kinds of wearable-sensors data. 2) there exists a large modality difference between wearable-sensor and vision-sensor in data dimension, data distribution, and inherent information content, which highlights the importance of adaptive feature fusion and specific knowledge transfer methods.

Based on these observations, in this paper, we focus on enhancing action recognition performance in vision-sensor modality (videos) by adaptively transferring the knowledge from multiple wearable-sensor modalities, meanwhile solving the aforementioned challenges. Since the knowledge distillation allows a model with only one modality input to achieve the performance close to the use of multiple modalities even with heterogeneous model and data \cite{hinton2015distilling,phuong2019towards}, we propose an end-to-end knowledge distillation framework, named Semantics-aware Adaptive Knowledge Distillation Network (SAKDN), which adaptively distills the complementary knowledge from multiple wearable-sensor modalities (teachers) to the vision-sensor modality (student), and concurrently improves the action recognition performance in vision-sensor modality (videos). An overview of the SAKDN is presented in Fig. \ref{Fig2}. In SAKDN, we use multiple kinds of wearable-sensor signals as teacher modalities and RGB stream of video as a single student modality. Since multi-modal action data share the same semantic content, we use semantics-aware knowledge of action class names to guide the multi-modal feature fusion, knowledge distillation, and representation learning.

More specifically, the SAKDN consists of multiple teacher networks and a single student network. The acceleration, gyroscope, and orientation signals are used as our teacher modalities, and the RGB videos are used as our student modality. To make the one-dimensional action data of wearable-sensor preserve the local temporal relationship and facilitate its visual recognition, we build a Gramian Angular Field (GAF) \cite{wang2015imaging} based virtual image generation model (as shown in Fig. \ref{Fig3}) which transforms the one-dimensional time-series signals into two-dimensional image representations and facilitates its application to existing visual models. Since there are multiple kinds of wearable-sensors modalities, we construct a Similarity-Preserving Adaptive Multi-modal Fusion Module (SPAMFM) to fully utilize the complementary information among different teacher networks. This module utilizes the intra-modality similarity, semantic embedding, and multiple relational knowledge to recalibrate the channel-wise features adaptively in each teacher network, as shown in Fig. \ref{Fig4}. To improve the performance of the student modality, we propose the Graph-guided Semantically Discriminative Mapping (GSDM) module, which transfers the graph-guided semantics-aware attention knowledge of multiple well-trained teacher networks to guide the training of the student network, as shown in Fig. \ref{Fig5}. Extensive experiments on three benchmarks verify that our SAKDN can realize adaptive knowledge transfer from multiple wearable-sensor modalities to vision-sensors modalities and achieve state-of-the-art performance.

The main contributions of this paper are as follows:

\begin{itemize}
\item To fully utilize the complementary knowledge from intermediate layers of multiple teacher networks, we propose a novel plug-and-play module, named Similarity-Preserving Adaptive Multi-modal Fusion Module (SPAMFM), which integrates intra-modality similarity, semantic embeddings, and multiple relational knowledge to learn the global context representation and recalibrate the channel-wise features adaptively in each teacher network.
\item To effectively exploit and transfer the knowledge of multiple well-trained teacher networks to the student network, we propose a novel knowledge distillation loss, named Graph-guided Semantically Discriminative Mapping (GSDM) module, which utilizes graph-guided ablation analysis to produce a good visual explanation highlighting the important regions in the image for predicting the semantic concept, and concurrently preserving respective interrelations of data for each modality.
\item One major advantage of our method is that it exploits the semantic relationship to bridge the modality gap between wearable-sensors and vision-sensors, and utilizes this constraint to guide the multi-modal feature fusion, knowledge transfer, and representation learning. The SAKDN focuses on the sensor-to-vision heterogenous action recognition problem and integrates SPAMFM, GSDM into a unified end-to-end adaptive knowledge distillation framework. Extensive experiments on three benchmark datasets validate the effectiveness of our SAKDN.
\end{itemize}

This paper is organized as follows: Section \uppercase\expandafter{\romannumeral2} briefly reviews the related works. Section \uppercase\expandafter{\romannumeral3} introduces the proposed SAKDN. Experimental results and related discussions are presented in Section \uppercase\expandafter{\romannumeral4}. Finally, Section \uppercase\expandafter{\romannumeral5} concludes the paper.

\section{Related Work}
\subsection{Uni-modal Action Recognition}
Action Recognition is an active research field and has received great attention in recent years \cite{aggarwal2011human}. The action recognition methods can be divided into three types: (1) handcrafted representations based \cite{chaquet2013survey,roshtkhari2013human,wang2016action,wang2016hierarchical,sargano2017comprehensive,ma2018space,siddiqui2018human,sargano2020human}, (2) graph learning based \cite{wang2013directed,mazari2019mlgcn,zhang2020semantics}, and (3) deep learning based \cite{ji20123d,simonyan2014two,tran2015learning,wang2016temporal,sargano2017human,lin2019tsm}. To be noticed, most of them are based on vision-sensors modality such as RGB, depth, skeleton, infrared images or videos, etc. Some representative RGB based works include IDT \cite{wang2013action}, DAG \cite{wang2013directed}, 3D CNN \cite{ji20123d}, two-stream CNNs \cite{simonyan2014two}, C3D \cite{tran2015learning}, TSN \cite{wang2016temporal}, TRN \cite{zhou2018temporal}, MLGCN \cite{mazari2019mlgcn}, TSM \cite{lin2019tsm}, etc. Yuan et al. \cite{yuan2016statistical} introduced a statistical hypothesis detector for abnormal RGB event detection in crowded scenes. Yuan et al. \cite{yuan2019memory} proposed a memory-augmented temporal dynamic learning network to learn temporal motion dynamics and tackle unsteady dynamics in long-duration motion of videos. Li et al. \cite{li2019deep}  introduced a spatio-temporal manifold network (STMN) that leverages data manifold structures to regularize deep action feature learning. In addition, other modalities (depth \cite{liu2019ntu,li2017visual}, skeleton \cite{yan2018spatial,meng2019sample}, infrared \cite{liu2018global}) based methods also receive increasing attention. Zhang et al. \cite{zhang2020semantics} proposed semantics-guided neural network (SGN) for skeleton-based action recognition. Zhang et al. \cite{zhang2017action} presented a low-cost descriptor called 3D histograms of texture (3DHoTs) to extract discriminant features from a sequence of depth maps. Though these vision-sensors based methods have achieved promising results, their performance is easily affected by camera viewpoints, background clutter, occlusion, and illumination change. In some privacy-sensitive area such as bank and government, the difficulty of acquiring images and videos make it infeasible. Furthermore, vision-based methods usually require expensive hardware resources to run deep learning models with high computational demands.

With the popularity of the wearable devices such as smartwatches and smartphones, human action recognition based on wearable-sensors has become a key research area in human activity understanding \cite{lara2012survey,wang2019deep}. Although aforementioned visions-sensors based methods have achieved good results, they cannot be directly applied to wearable-sensor based problems due to the existence of huge modality divergence. Since wearable-sensors action data is suitable for privacy protecting and robust to variant illuminations and camera viewpoints, some specific works have been proposed recently. Jiang et al. \cite{jiang2015human} assembled signal sequences of accelerators and gyroscopes into an activity image to learn optimal representations automatically. Wannenburg et al. \cite{wannenburg2016physical} utilized ten different classifier algorithms to classify the human actions using the accelerator signals captured by smartphones. Setiawan \cite{setiawan2019deep} used gramian angular field to transform one-dimensional wearable-sensor signals to two-dimensional images. Wang et al. \cite{wang2019attention} proposed an attention-based CNN framework to address weakly-supervised sensors-based action recognition problem. Fazli et al. \cite{fazli2020hhar} built a hierarchical classification with neural networks to recognize human activities based on built-in sensors in smart and wearable devices. Different from vision-sensor based methods, most of these wearable-sensor based action recognition methods are based on raw sensor time-series signals, which lack color and texture information and could not preserve the local temporal relationship. In addition, these methods use simple feature fusion methods to fuse the knowledge from different sensor modalities without considering intra-modality similarity, semantic embeddings, and multiple relational knowledge.

To increase the representative ability of wearable-sensors action features, we construct a Gramian Angular Field (GAF) based virtual image generation model, which transforms the one-dimensional time-series signals of wearable-sensors into two-dimensional image representations. To fully utilize the complementary knowledge from multiple wearable-sensors, we propose a Similarity- Preserving Adaptive Multi-modal Fusion Module (SPAMFM), which integrates intra-modality similarity, semantic embeddings, and multiple relational knowledge to learn the global context representation and recalibrate the channel-wise features adaptively.

\subsection{Multi-modal Action Recognition}
Action recognition has been developed for a long period, but action recognition on multiple modalities is a relatively new topic. With the development of deep learning methods and various hardware such as cameras and wearable devices, there are some typical methods of dealing with multi-modal action recognition problems in recent years. These methods can be roughly categorized into three types: 1) cross-view action recognition,  typical works \cite{liu2018hierarchically,wang2019generative} used transfer learning methods to reduce the domain gap of action data from different camera views; 2) cross-spectral action recognition, typical works \cite{shahroudy2017deep,liu2018transferable} addressed the visible-to-infrared action recognition problems using domain adaptation methods. Yuan et al. \cite{yuan2018action} proposed a spatial-optical data organization and sequential learning framework based on spatial-optical action data; 3) cross-media action recognition, typical works \cite{yu2019exploiting,8931264} designed specific multi-modal feature learning frameworks to address the image-to-video action recognition problems.

Different from these cross-domain action recognition problems, the multi-modal action recognition based on wearable-sensors and vision-sensors (sensor-to-vision) is essentially a heterogeneous knowledge transfer problem because there exists large modality difference between wearable-sensors and vision-sensors in data dimension, data distribution, and inherent information content, which is shown in Fig. \ref{Fig1}. And related research about sensor-to-vision action recognition is limited. Chen et al. \cite{chen2014improving} proposed a feature fusion framework to combine signals from the depth camera and inertial body sensor. Kong et al. \cite{kong2019mmact} built a multi-modality distillation model with the attention mechanism to realize adaptive knowledge transfer from sensor modalities to vision modalities. Hamid et al. \cite{joze2019mmtm} proposed a multi-modal transfer module to fuse knowledge from different unimodal CNNs and tested this module for three different multi-modal fusion tasks: gesture recognition, audio-visual speech enhancement, and action recognition. However, most of these methods only use raw one-dimensional time-series sensor signals to recognize actions. Since time-series data lacks local temporal relationship, color, and texture information, it may affect the representative ability of wearable-sensor signals and make existing pre-trained deep learning models (e.g. LeNet, AlexNet, VGGNet, ResNet, etc.)  hard to adapt. Furthermore, the semantic relationship between wearable-sensors and vision-sensors action data is ignored in previous works, which can guide the knowledge transfer.

In this paper, we use the semantics-aware information to guide the multi-modal feature fusion, knowledge distillation, and representation learning of our SAKDN. Although method \cite{joze2019mmtm} built a squeeze and excitation based multi-modal feature fusion module which only used the simple concatenation of features from different modalities to learn the global context embedding without considering more diverse relation functions and the intra-modality similarity relationship. Differently, we propose a novel plug-and-play module, named Similarity-Preserving Adaptive Multi-modal Fusion Module (SPAMFM), which seamlessly integrates intra-modality similarity, semantic embeddings, and multiple relational knowledge to learn the global context representation and recalibrate the channel-wise features adaptively in each network.

\subsection{Knowledge Distillation}

Knowledge distillation is a general technique for supervising the training of student networks by capturing and transferring useful knowledge from well-trained teacher networks. Hinton et al. \cite{hinton2015distilling} used softened labels of the teacher with a temperature to transfer knowledge to a small student network. Attention transfer \cite{DBLP:conf/iclr/ZagoruykoK17} designed a knowledge distillation loss based on summed $p$-norm of convolutional feature activations along channel dimension. Park et al. \cite{park2019relational} proposed distance-wise and angle-wise distillation losses to realize relational knowledge transfer. Tung et al. \cite{tung2019similarity} construct a knowledge distillation loss with the constraint that input pairs that produce similar (dissimilar) activations in the teacher network should produce similar (dissimilar) activations in the student network. Hoffman et al. \cite{hoffman2016learning} built a modality hallucination architecture for training an RGB object detection model using depth as side information. Garcia et al. \cite{garcia2018modality} proposed a generalized distillation framework considering the case of learning representations from the depth and RGB videos, while relying on RGB data only at test time. Crasto et al. \cite{crasto2019mars} introduced a feature-based loss compared to the Flow stream, a linear combination of the feature-based loss and the standard cross-entropy loss, to mimic the motion stream, and as a result avoids flow computation at test time.

Different from existing knowledge distillation methods that focus on the modality transfer task across vision-sensor based modalities, we move a further step towards knowledge transfer from wearable-sensor based modalities to vision-sensors based modalities. In this paper, we construct a novel knowledge distillation module, named Graph-guided Semantically Discriminative Mapping (GSDM), which utilizes graph-guided ablation analysis to produce good visual explanations highlighting the important regions for predicting the semantic concept and preserving the intrinsic structures concurrently. Since the semantic relationship between wearable-sensors and vision-sensors is similar, we transfer the semantics-aware attention knowledge of multiple well-trained teacher networks to guide the training of the student network.

\section{Semantics-aware Adaptive Knowledge Distillation Networks}
\subsection{Framework Overview}
The framework of the SAKDN is shown in Fig. \ref{Fig2}, which is an end-to-end knowledge distillation framework seamlessly constituted by three parts: virtual image generation of wearable-sensors, training of multiple teacher networks, and multi-modality knowledge distillation from multiple teacher networks to the student network. We use wearable-sensors action data (acceleration, gyroscope, and orientation) as teacher modalities and RGB videos as student modalities. The virtual image generation part uses Gramian Angular Field (GAF) \cite{wang2015imaging,setiawan2019deep} to encode one-dimensional time-series signals of wearable-sensor into the two-dimensional image representation. We build a novel Similarity-Preserving Adaptive Multi-modal Fusion Module (SPAMFM) to fuse intermediate representation knowledge from different teacher networks adaptively. Then we use semantic preserving loss along with cross-entropy loss for the training of multiple teacher networks. The multi-modality knowledge distillation consists of two knowledge distillation losses, one is our proposed Graph-guided Semantically Discriminative Mapping (GSDM) loss, and the other one is the soft-target knowledge distillation loss \cite{hinton2015distilling}. We train the student network using GSDM loss, soft-target loss, semantic preserving loss, and cross-entropy loss.

\subsection{Virtual Image Generation}
\begin{figure}[!t]
\centering
\includegraphics[scale=0.23]{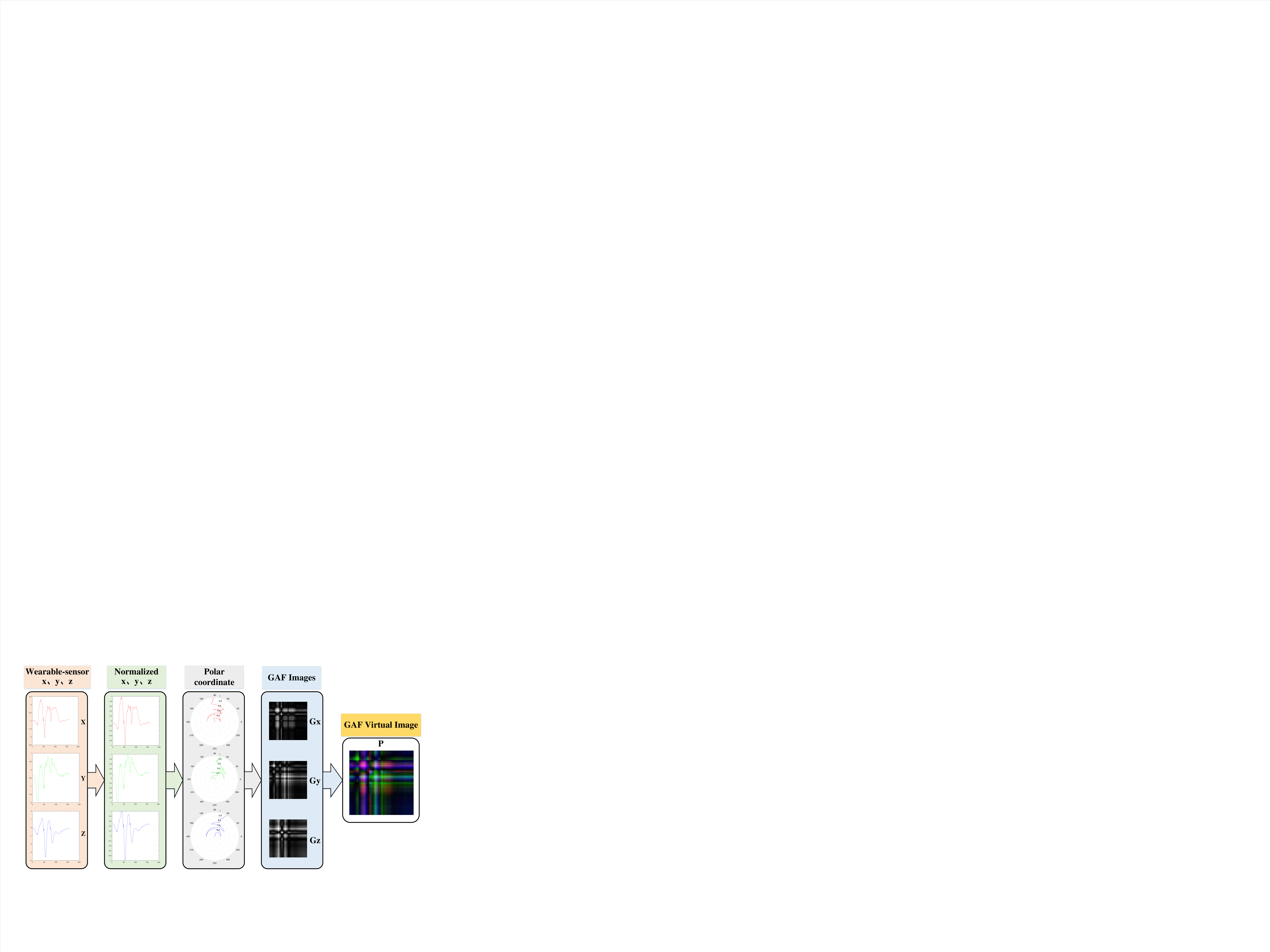}
\caption{Overview of Gramian Angular Field (GAF) based virtual image generation framework. }
\vspace{-10pt}
\label{Fig3}
\end{figure}

To make the one-dimensional action data of wearable-sensor preserve the local temporal relationship and facilitate its visual recognition, we build a Gramian Angular Field (GAF) based virtual image generation model which transforms the one-dimensional time-series signals into two-dimensional image representations. The Gramian Angular Field (GAF) based virtual image generation model is shown in Fig. \ref{Fig3}. Since there are three axial time-series signals (x, y, z) of wearable-sensors action data, we denote one of the tri-axial signals as $X=\{x_1, \cdots, x_n\}$. We then use min-max normalization to normalize original signal $X$  into the interval $[-1,1]$ and get normalized signal $\widetilde{X}$,
\begin{equation}\label{eq1}
\widetilde{X}_i=\frac{(x_i-\textrm{max}(X))+(x_i-\textrm{min}(X))}{\textrm{max}(X)-\textrm{min}(X)}
\end{equation}

Then, we use transformation function $g$ to transform the normalized signal $\widetilde{X}$ to the polar coordinate system, which represents cosine angle from the normalized amplitude and the radius from the time $t$, as represented in Eq. (\ref{eq2}).
\begin{equation}\label{eq2}
g(\widetilde{x}_i,t_i)=[\theta_i,r_i]  ~~\textrm{where}~~ \left\{
\begin{aligned}
\theta_i & =  \arccos(\widetilde{x}_i), \widetilde{x}_i\in \widetilde{X}\\
r_i& =  t_i
\end{aligned}
\right.
\end{equation}

After encoding the normalized time-series signals into a polar coordinate system, the correlation coefficient between time intervals can be easily obtained by trigonometric sum between points. Since the correlation coefficient can be calculated by the cosine of the angle between vectors \cite{wang2015imaging,setiawan2019deep}, the correlation between time $i$ and $j$ is calculated using $\cos(\theta_i+\theta_j)$ and the Gramian Angular Field based matrix is defined as $G$:
\begin{equation}\label{eq3}
G=
 \left(
   \begin{array}{ccc}
     \textrm{cos}(\theta_1+\theta_1) &\cdots& \textrm{cos}(\theta_1+\theta_n)  \\
     \vdots & \ddots & \vdots \\
     \textrm{cos}(\theta_n+\theta_1) & \cdots &\textrm{cos}(\theta_n+\theta_n) \\
   \end{array}
 \right)
\end{equation}

The encoding map of Eq. (\ref{eq3}) has two important properties. Firstly, it is bijective as cos$(\phi)$ is monotonic when $\phi\in[0,\pi]$. Given the time-series data, the proposed map produces one and only one result in the polar coordinate system with a unique inverse map. Secondly, as opposed to Cartesian coordinates, polar coordinates preserve absolute temporal relations. After encoding the scaled time-series signals into a polar coordinate system, we can easily extract the correlation coefficient between time intervals considering the trigonometric sum between points. In this way, the GAF provides a new representation style that can preserve the local temporal relationship in the form of temporal correlation as the timestamp increases. For wearable-sensor based action data, the accelerator, gyroscope, and orientation signals are in tri-axial style. Therefore, we assume that each axis sensor data with length $n$ can be transformed into a single GAF matrix with the size $n\times n$. Then, the GAF matrices of tri-axial sensor data (x-, y-, and z-axis) are assembled as a three-channel image representation $P=\{G_x,G_y,G_z\}$ of size $n \times n \times 3$. And this novel image representation is named as GAF based Virtual Image (GAFVI). GAFVI of wearable-sensors will be used as the input for teacher modalities in this paper.

\subsection{Similarity-Preserving Adaptive Multi-modal Fusion}

\begin{figure}[!t]
\centering
\includegraphics[scale=0.2]{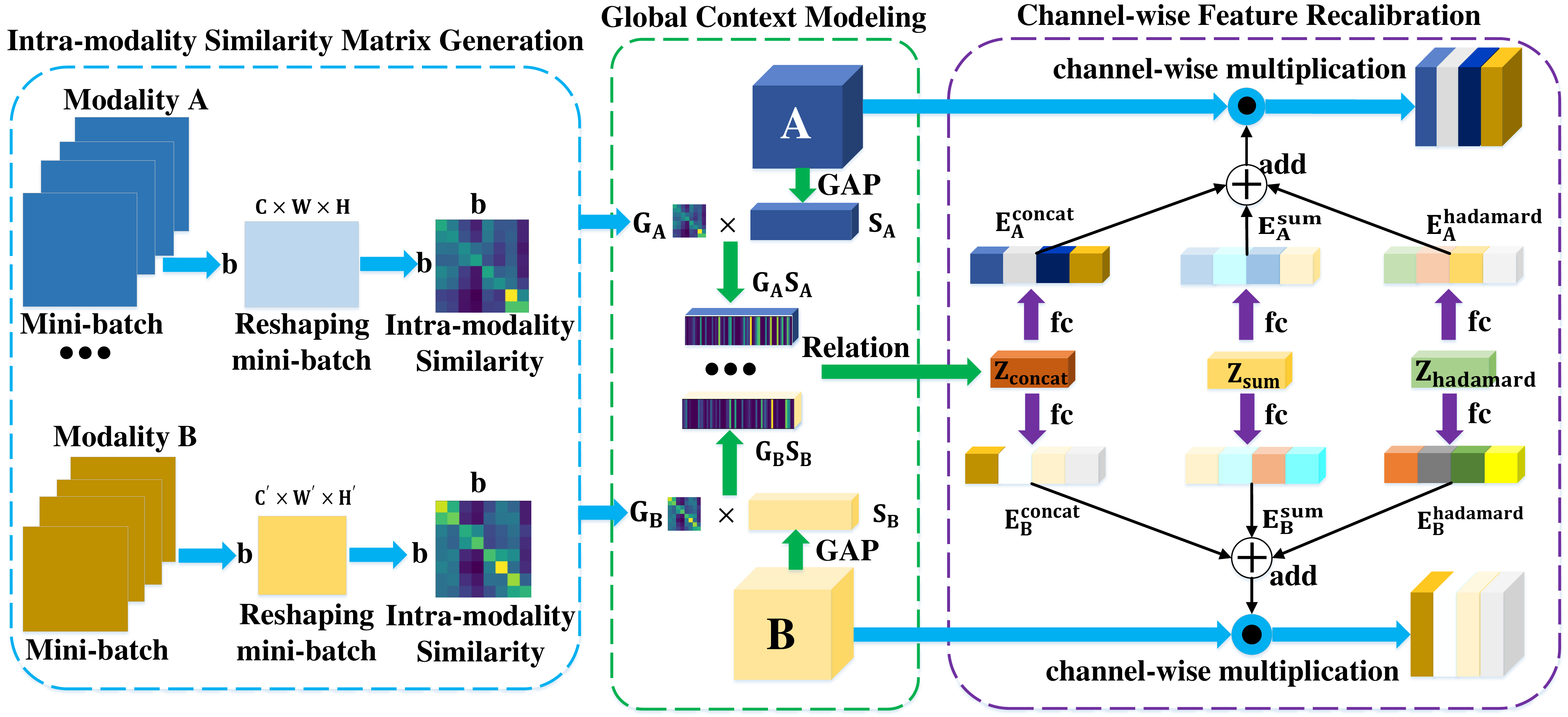}
\caption{Architecture of SPAMFM for two modalities. $A$ and $B$ denote the features at a given layer of two CNNs. }
\vspace{-10pt}
\label{Fig4}
\end{figure}

Since there exist multiple teacher modalities when training the teacher networks, we need to discover and fuse representative features among these modalities. To be noticed, attention gives a feasible way to extract important features for prediction. However, most of the existing attention operations \cite{DBLP:conf/iclr/ZagoruykoK17,hu2018squeeze,zhao2020exploring} focus on the uni-modal problem and neglect the relationship knowledge among different modalities when conducting multi-modal fusion. To fully utilize the complementary knowledge from multiple teacher modalities, we extend the attention operations from uni-modal to multi-modal feature fusion by integrating the intra-modality similarity, semantic embeddings, and multiple relational knowledge into a unified Similarity-Preserving Adaptive Multi-modal Fusion (SPAMFM) module. The simplest case of SPAMFM for two modalities is shown in as shown in Fig. \ref{Fig4}.

\subsubsection{Intra-modality Similarity Matrix Generation}
Assume that we have $m$ teacher modalities and each modality has its own network $\{T_k| k=1,\cdots, m\}$. Given an input mini-batch of size $b$, the activation map produced by the teacher network $T_k$ at a particular layer $l$ is denoted as $A_{T_k}^{l}\in \mathbb{R}^{b\times c_k\times h_k \times w_k}$, where $b$ is the batch size, $c_k$ is the number of output channels for the $k$-th modality, and $h_k$, $w_k$ are spatial dimensions. Inspired by attention-based knowledge transfer methods \cite{DBLP:conf/iclr/ZagoruykoK17,tung2019similarity} which use activation correlation to conduct knowledge transfer, we use mini-batch data to calculate intra-modality similarities in particular intermediate layers for different teacher modalities. Specifically, the activation maps $A_{T_k}^{l}$ are first reshaped to $R_{T_k}^{l}\in \mathbb{R}^{b\times c_kh_kw_k}$, and then we use row-wise L2-normalized outer product of $R_{T_k}^{l}$ matrices to calculate intra-modality similarity-preserving matrices $G_{T_k}^{l}\in \mathbb{R}^{b\times b}$:
\begin{equation}\label{eq4}
\tilde{R}_{T_k}^{l}=R_{T_k}^{l}\times R_{T_k}^{l\top}\\
\end{equation}
\begin{equation}\label{eq5}
G_{T_k[i,:]}^{l}=\frac{\tilde{R}_{T_k[i,:]}^{l}}{\left\|\tilde{R}_{T_k[i,:]}^{l}\right\|_2}
\end{equation}
where $\tilde{R}_{T_k}^{l}$ encodes the similarity of the activations within teacher modality $k$ of layer $l$ in the mini-batch, $[i, :]$ denotes the $i$-th row in a matrix. These intra-modality similarities $G_{T_k[i,:]}^{l}$ can be utilized as weight matrices to guide the fusing of different relation functions in global context modeling. In this way, the calculated global context information can adaptively preserve the intra-modality relationship as well as the complementary information among different teachers.

\subsubsection{Global Context Modeling and Feature Recalibration}

After obtaining the intra-modality similarity matrices $\{G_{T_k}^{l}| k=1, \cdots, m\}$, we build a global context modeling module to receive features from particular layers (conv or fc) of different teacher networks and learns a similarity-preserved global context embedding, then we use this embedding to recalibrate the input features from different modalities. To fix notation, we let $A_k^l\in \mathbb{R}^{ b\times c_k\times h_k\times w_k}$ denotes the feature maps of a batch at a given layer $l$ of modality $k$. We first use global average pooling (GAP) to generate squeezed feature vectors $S_k^l\in\mathbb{R}^{ b\times c_k}$ for different modalities. Formally, a statistic $S_k^l$ is generated by shrinking $A_k^l$ through spatial dimensions $h_k\times w_k$, where the $c$-th element of $S_k^l$ is calculated by:
\begin{equation}\label{eq6}
S_k^l(b,c)=\frac{1}{h_k\times w_k}\sum_{i=1}^{h_k}\sum_{j=1}^{w_k}A_k^l(b,c,i,j)
\end{equation}

In order to make the global context preserve the intra-modality relationship as well as the complementary information among different teacher modalities, we use the product of intra-modality similarity matrix $G_{T_k}^{l}$ and
squeezed feature vector $S_k^l$ for each teacher modality to learn joint representation. To aggregate their complementary heterogeneous information from different aspects, we use three different relation functions: concatenation,
summation, and Hadamard product, which have been validated their effectiveness in \cite{zhao2020exploring}. Thus, we can get three forms of joint representations through three independent fully-connected layers with three kinds of relation functions:
\begin{equation}\label{eq7}
Z_{con}^l=W_{con1}^l[G_{T_1}^{l}S_1^l, \cdots, G_{T_m}^{l}S_m^l]+b_{con1}^l
\end{equation}
\begin{equation}\label{eq8}
Z_{sum}^l=W_{sum1}^l\left(\sum_{k=1}^m{G_{T_k}^{l}S_k^l}\right)+b_{sum1}^l
\end{equation}
\begin{equation}\label{eq9}
Z_{had}^l=W_{had1}^l\prod_{k=1}^m{G_{T_k}^{l}S_k^l}+b_{had1}^l
\end{equation}
where $[\cdot,\cdot]$ denotes the concatenation operation, $\prod_{k=1}^m$ denotes hadamard product from modality $1$ to modality $m$, $Z_{con}^l\in \mathbb{R}^{c_{con}}$, $Z_{sum}^l\in \mathbb{R}^{c_{sum}}$ and $Z_{had}^l\in \mathbb{R}^{c_{had}}$ denote joint representations of $l$-th layer for concatenation, summation and hadamard product relation functions, respectively. Here, $W_{con1}^l\in \mathbb{R}^{c_{con}\times \sum_{k=1}^{m}{c_k}}$, $W_{sum1}^l\in \mathbb{R}^{c_{sum}\times c_k}$, $W_{had1}^l\in \mathbb{R}^{c_{had}\times c_k}$ are weights, $b_{con1}^l\in \mathbb{R}^{c_{con}}$, $b_{sum1}^l\in \mathbb{R}^{c_{sum}}$ and $b_{had1}^l\in \mathbb{R}^{c_{had}}$ are the biases of the fully-connected layers. We choose  $c_{con}=\frac{\sum_{k=1}^{m}{c_k}}{2m}$, $c_{sum}=c_k$ and $c_{had}=c_k$ according to \cite{hu2018squeeze} to restrict the model capacity and increase its generalization ability.

To make use of the global context information aggregated in the above three joint representations $Z_{con}^l$, $Z_{sum}^l$ and $Z_{had}^l$, we predict excitation signals for them through three independent fully-connected layers:
\begin{equation}\label{eq10}
E_{con}^l=W_{con2}^{l}Z_{con}^l+b_{con2}^l
\end{equation}
\begin{equation}\label{eq11}
E_{sum}^l=W_{sum2}^lZ_{sum}^l+b_{sum2}^l
\end{equation}
\begin{equation}\label{eq12}
E_{had}^l=W_{had2}^lZ_{had}^l+b_{had2}^l
\end{equation}
where $W_{con2}^l\in \mathbb{R}^{c_k \times c_{con}}$, $W_{sum2}^l\in \mathbb{R}^{c_k\times c_{sum}}$, $W_{had2}^l\in \mathbb{R}^{c_k\times c_{had}}$ are weights, $b_{con2}^l\in \mathbb{R}^{c_k}$, $b_{sum2}^l\in \mathbb{R}^{c_k}$ and $b_{had2}^l\in \mathbb{R}^{c_k}$ are the biases of the fully connected layers.

After obtaining these three excitation signals $E_{con}^l\in \mathbb{R}^c_k$, $E_{sum}^l\in \mathbb{R}^c_k$ and $E_{had}^l\in \mathbb{R}^c_k$, we use them to recalibrate the input feature $A_k^l$  from each modality $k$ adaptively by a simple gating mechanism,
\begin{equation}\label{eq13}
\tilde{A}_k^l=(\delta(E_{con}^l)+\delta(E_{sum}^l)+\delta(E_{had}^l))\odot {A}_k^l
\end{equation}
where $\odot$ is channel-wise product operation for each element in the channel dimension, and $\delta(\cdot)$ is the ReLU function. With SPAMFM, we can realize adaptive multi-modal feature fusion and inter-modality feature recalibration, which allows the features of one modality to recalibrate the features of another modality while concurrently preserving the intra-modality similarities as well as the complementary information among different teacher modalities.

\subsection{Graph-guided Semantically Discriminative Mapping}

Traditional knowledge distillation methods usually conduct knowledge transfer at the last fully-connected layers and ignore the intermediate layers which contain abundant essential complementary information between networks. For heterogeneous knowledge problems like sensor-to-vision action recognition, the knowledge in intermediate layers is important for efficient knowledge transfer. To mitigate the modality divergence between teacher and student modalities, we propose a novel semantics-aware knowledge distillation module, named Graph-guided Semantically Discriminative Mapping (GSDM), which works at convolutional layers and transfers the semantics-aware attention knowledge of multiple well-trained teacher networks to guide the training of student network. This module utilizes graph-guided ablation analysis to produce good visual explanations for both teacher and student modalities highlighting the important regions for predicting the semantic concept and concurrently preserving respective
interrelations of data for each modality.

\begin{figure}[!t]
\centering
\includegraphics[scale=0.38]{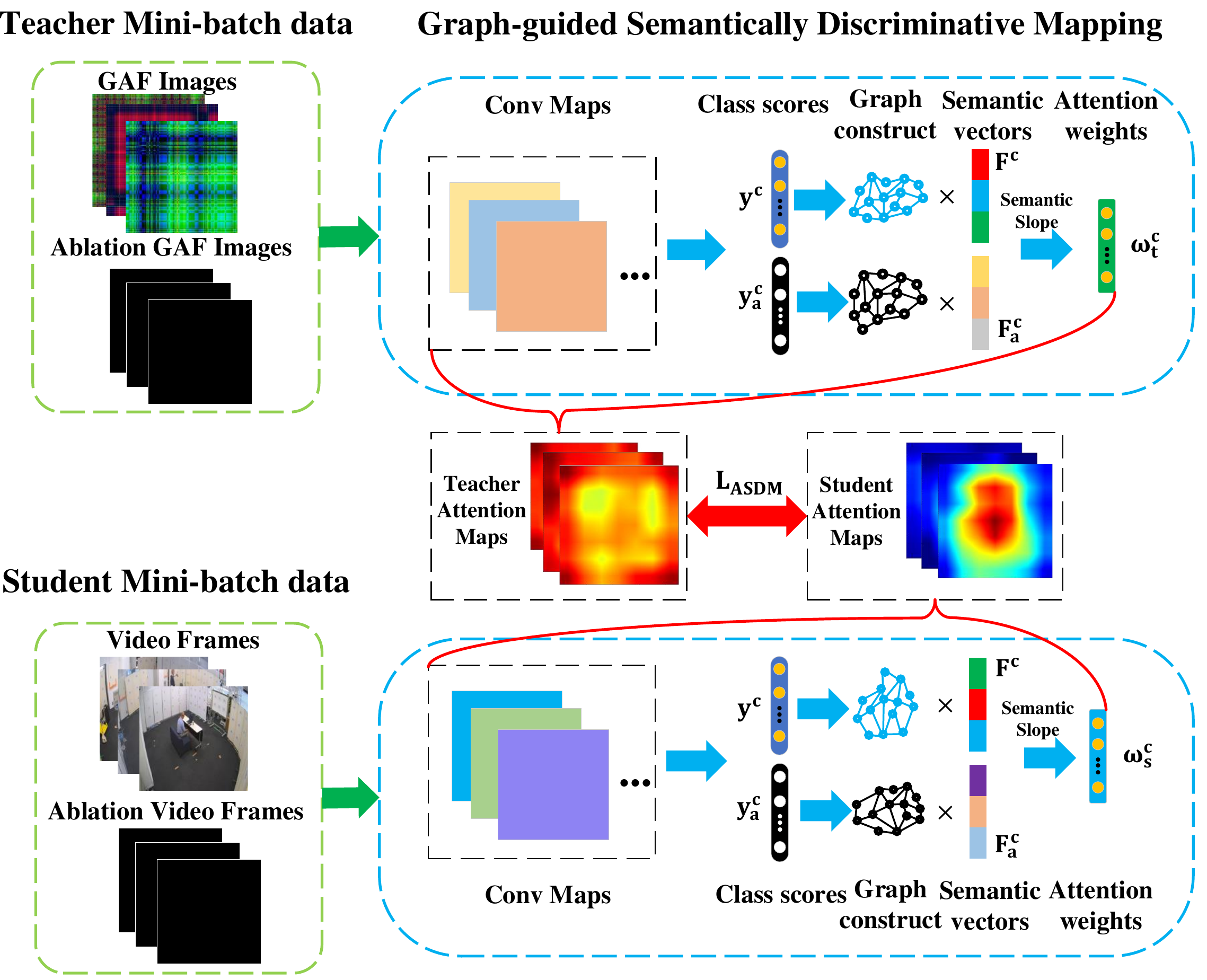}
\caption{Overview of Graph-guided Semantically Discriminative Mapping (GSDM) knowledge distillation module.}
\vspace{-10pt}
\label{Fig5}
\end{figure}

Since previous works \cite{DBLP:conf/iclr/MorcosBRB18,DBLP:journals/corr/abs-1806-02891,desai2020ablation} have validated that the ablation of some units of a network can be an indicator of how important a unit is for a particular class, we use ablation drop of mini-batch input features to produce visual explanations based knowledge distillation loss across domains. Different from previous methods which use global average pooled gradients and class scores for visual explanation, we use semantics-guided ablation analysis to learn the visual explanations because the similar semantic relationship between wearable-sensors and vision-sensors data can be considered as good guidance for knowledge transfer while class scores are too strict for heterogenous sensor-to-vision action recognition problem. The framework of GSDM is shown in Fig. \ref{Fig5}.

The input mini-batch data of student network contains two parts, the first part $I\in \mathbb{R}^{b\times c\times h \times w}$ contains raw input mini-batch data, the second part $I_a\in \mathbb{R}^{b\times c\times h \times w}$ contains black images which is essentially the ablation of the raw input data. And the combination of these two parts $[I;I_a]\in \mathbb{R}^{2b\times c\times h \times w}$ is used as the input. We assume that the class score $y^c$ for class $c$ can be considered as a non-linear function of input data. When we set all the input mini-batch data to zeros and repeat the forward pass, we get a reduced activation score $y_a^c$ with respect to feature map $A_p$ of $p$-th unit. Based on these class scores $y^c$ and $y_a^c$, we use Glove \cite{pennington2014glove} to calculate their corresponding semantic embeddings $F^c$ and $F_a^c$:
\begin{equation}\label{eq14}
F=\left\{
    \begin{array}{ll}
      F^c=\textrm{Glove}(y^c)\in \mathbb{R}^{b\times 300},  \hbox{when input is $I$;} \\
      F_a^c=\textrm{Glove}(y_a^c)\in \mathbb{R}^{b\times 300},  \hbox{when input is $I_a$.}
    \end{array}
  \right.
\end{equation}

Since manifold learning \cite{nie2010flexible} extracts intrinsic structures from data, we construct two graphs for original data $I$ and ablation data $I_a$, where vertexes are embedded features at the final fully connected layers and edges are the relations between features. The edge weights $W_{i,j}$ between the input data $x_i$ and $x_j$ are determined by the Gaussian similarity, $W_{i,j}=\textrm{exp}(-\frac{\|f_i-f_j\|^2}{2})$, where $f_i$ and $f_j$ are embedded feature vectors of $x_i$ and $x_j$. Then we apply the normalized graph Laplacians \cite{chung1997spectral} on $W$, that is, $Q=D^{1/2}WD^{-1/2}$, where $D$ is a diagonal matrix with its $(i, i)$-value to be the sum of the $i$-th row of $W$. In this way, the manifold structure in the data can be well represented in graph matrix $Q\in \mathbb{R}^{b\times b}$.

To preserve the embedded low-dimensional manifold subspace structure of original modalities when conducting knowledge distillation, we first multiply the semantic embeddings with the graph matrix, which is essentially the manifold regularization. In this way, a semantic space that is robust against small perturbations can be produced \cite{zhou2004learning}. Semantics propagation can be seen as a repeated random walk through the graph of features using an affinity matrix to assign the semantic embeddings of data, which can effectively preserves the manifold structure \cite{DBLP:conf/iclr/LiuLPKYHY19}. Then, we define a graph-guided slope metric $\omega_{p,l}^c\in \mathbb{R}^{b\times 300}$ to measure the changing rate of the transformed semantic embeddings for the p-th unit of layer $l$ and class $c$.

\begin{equation}\label{eq15}
 \omega_{p,l}^c=\frac{QF^c-Q_aF_a^c}{QF^c}
\end{equation}
where $Q\in \mathbb{R}^{b\times b}$ and $Q_a\in \mathbb{R}^{b\times b}$ are the normalized graph similarity matrices for original data and ablation data, respectively. In this way, the intrinsic of data can be preserved and concurrently the importance value can be represented by the fraction of drop in semantic embeddings of class $c$ when the input features are removed. Then the graph-guided semantically discriminative map $M_l^c$ for the $l$-th layer of class $c$ can be obtained as a weighted linear combination of activation maps $A_{p,l}$ and corresponding weights $\omega_{p,l}^c$,
\begin{equation}\label{eq16}
M_{p,l}^c=\textrm{ReLU}\left(\sum_p \omega_{p,l}^cA_{p,l}\right)
\end{equation}
The dimensionality of the weight $\omega_{p,l}^c$ is adaptively adjusted to the dimensionality of different feature maps in the same way as the Grad-CAM \cite{selvaraju2017grad}. The GSDM of specific layers for teacher and student networks are generated following Eq. (\ref{eq16}). After obtaining the GSDM, the GSDM based knowledge distillation loss can be constructed. Assume that we have $m$ teacher modalities and one student modality, we use mean squared error (MSE) loss between the normalized GSDM of teachers and student to transfer knowledge:
\begin{equation}\label{eq17}
L_{\textrm{GSDM}}=\frac{\sum_{k=1}^{m}\sum_{l_T\in \mathcal{L}_{distill}^T,l_S\in \mathcal{L}_{distill}^S}\left\|\frac{M_{l_T}^{T_k}}{\|M_{l_T}^{T_k}\|_2}-\frac{M_{l_S}^{S}}{\|M_{l_S}^{S}\|_2}\right\|_2^2}{m\times N^{\mathcal{L}_{distill}^T}}
\end{equation}
where $M_{l_T}^{T_k}$ denotes the GSDM for teacher network $T_k$ of the $l$-th layer, $M_{l_S}^S$ is the GSDM for student network of the $l$-th layer, $ \mathcal{L}_{distill}^T$ represents the group containing the choosing layers of teacher networks for knowledge distillation, $ \mathcal{L}_{distill}^S$ is the group containing the choosing layers of student network, and $\|\cdot\|_2$ denotes the L2 norm. $N^{\mathcal{L}_{distill}^T}$ is the number of choosing layers in group $ \mathcal{L}_{distill}^T$.

\subsection{Semantics-aware Adaptive Knowledge Distillation}

Based on the Virtual Image Generation model, Similarity-Preserving Adaptive Multi-modal Fusion Module (SPAMFM), and Graph-guided Semantically Discriminative Mapping (GSDM) knowledge distillation module, we seamlessly integrate them into a unified adaptive knowledge distillation framework, named Semantics-aware Adaptive Knowledge Distillation Network (SAKDN), to address the sensor-to-vision heterogenous action recognition problem, shown in Fig. \ref{Fig2}. In SAKDN, we use multiple wearable-sensors as teacher modalities and use vision-sensor as student modalities. For teacher networks, the input is GAF images of wearable-sensors. And the input of the student network is the RGB video.

\subsubsection{Training of teacher networks}
Assume that we have $m$ wearable-sensors modalities, we build $m$ teacher networks using VGG16 \cite{simonyan2014very} as the backbone. As shown in Fig. \ref{Fig2}, given GAF images for each modality, we simultaneously feed them into their respective networks for model training. The SPAMFM are added into selected layers of VGG16 among teacher networks, which is shown as follows:
\begin{equation}\label{eq18}
\mathcal{L}_{\textrm{SPAMFM}}=\{\textrm{conv}_1^2, \textrm{conv}_2^2, \textrm{conv}_3^3, \textrm{conv}_4^3, \textrm{conv}_5^3, \textrm{fc}1, \textrm{fc}2\}
\end{equation}

In addition to the SPAMFM, we design a semantic preserving loss at the fc2 layer among different teacher networks to make sure the fc2 layer contain semantic knowledge as well as the intra-modality relationship and the complementary information from different teacher modalities. The semantic preserving loss $L_{\textrm{SP}}^T$ is defined as the MSE loss between the raw features of the fc2 layer and their corresponding semantic representations of action class names,
\begin{equation}\label{eq19}
L_{\textrm{SP}}^T=\frac{1}{m}\sum_{k=1}^m\left\|H_k^T-F_k\right\|_2^2
\end{equation}
where $H_k^T$ is the raw feature of the fc2 layer for teacher network $k$, $F_k$ is the corresponding semantic representation.

All the teacher networks are trained simultaneously using $L_{\textrm{SP}}^T$ along with the summation of cross-entropy loss for all teacher networks $L_{\textrm{CS}}^T$. The total loss $L_{\textrm{T}}$ for all teacher networks is organized as:
\begin{equation}\label{eq20}
L_{\textrm{T}}= L_{\textrm{CS}}^T+L_{\textrm{SP}}^T=\frac{1}{m}\sum_{k=1}^m\textrm{CE}(Y_k^T,Z_k^T)+\frac{1}{m}\sum_{k=1}^m\left\|H_k^T-F_k\right\|_2^2
\end{equation}
where CE is the cross entropy, $Y_k$ and $Z_k$ denote the predicted labels and class probability for teacher network $k$, respectively.

\subsubsection{Training of student network}
Our student network is a TRN \cite{zhou2018temporal} with BN-Inception using only RGB videos as input. During the training of the student network, the parameters of teacher networks are fixed, as shown in Fig. \ref{Fig2}. In order to reduce the computational cost during the training phase, we only perform distillation on some representative features. Thus, the GSDM is added into selected convolutional layers between BN-Inception and VGG16 networks for all teacher-student pairs, which are shown as follows:
\begin{equation}\label{eq21}
\mathcal{L}_{distill}^T=\{\textrm{conv}_1^2, \textrm{conv}_2^2, \textrm{conv}_3^3, \textrm{conv}_4^3,  \textrm{conv}_5^3\}
\end{equation}
\begin{equation}\label{eq22}
\mathcal{L}_{distill}^S=\{\textrm{conv}2, \textrm{Inc}{3c}, \textrm{Inc}{4c},  \textrm{Inc}{5a}, \textrm{Inc}{5b}\}
\end{equation}
where $\textrm{conv}_i^j$ represents the $j$-th convolutional activation map of convolution group $i$, Inc represents the inception layer.

In addition to the GSDM distillation loss $L_{\textrm{GSDM}}$ in Eq. (\ref{eq17}), we build a complementary knowledge distillation loss at the last fully-connected layers between teacher and student networks.
\begin{equation}\label{eq23}
L_{\textrm{ST}}=\frac{1}{m}\sum_{k=1}^{m}\textrm{KL}(\frac{P_k^T}{T},\frac{P^S}{T})
\end{equation}
where $\textrm{KL}(\cdot,\cdot)$ is the Kullback-Leibler divergence, $P_k^T$ is the class probability prediction of teacher network $k$, $P^S$ is the class probability prediction of the student network, $T$ denotes the temperature controlling the distribution of the probability. We set $T=4$ in this paper suggested by \cite{hinton2015distilling}.

To make the semantic knowledge between teacher and student networks similar, we use semantic preserving loss between the fc1 layer of the student network and the fc2 layer of teacher networks, which is defined as follows:
\begin{equation}\label{eq24}
L_{\textrm{SP}}^S=\frac{1}{m}\sum_{k=1}^{m}\left\|H^S-H_k^T\right\|_2^2
\end{equation}
where $H^S$ denotes the features of the fc1 layer for the student network,  $H_k^T$ represents the features of the fc2 layer for teacher network $k$. Since we use Eq. (\ref{eq19}) to train the teacher networks, the fc2 layer of the trained teacher network already contains semantic knowledge. Therefore, we can realize semantic knowledge preserving for student network using Eq. (\ref{eq24}).

To train the student network, we use cross entropy loss $L_{\textrm{CS}}^S=CE(Y^S,Z^S)$ along with two knowledge distillation loss $L_{\textrm{GSDM}}$ and $L_{\textrm{ST}}$, and semantic preserving loss $L_{\textrm{SP}}^S$. The total loss $L_S$ for student network is defined as follows:
\begin{equation}\label{eq25}
L_S=L_{\textrm{CS}}^S+\alpha L_{\textrm{ST}}+\beta L_{\textrm{GSDM}}+\gamma L_{\textrm{SP}}^S
\end{equation}
where $Y^S$ and $Z^S$ are predicted labels and class probability for the student network, respectively. $\alpha$, $\beta$, and $\gamma$ are the parameters controlling the importance of  ST, GSDM, and SP loss, respectively.

\section{Experiments}
\subsection{Experimental Setup}

In this work, we conduct extensive experiments on three benchmarks for sensor-to-vision action recognition. We first introduce the three datasets and implementation details. Then, we compare our SAKDN with existing knowledge distillation and action recognition methods. In addition, we conduct ablation studies to analyze the importance of the proposed ST, SPAMFM, GSDM, and SP. Furthermore, we conduct experiments with different backbone architectures and selected transfer layers to validate whether our SAKDN could generalize to different networks and choosing layers. To show how we select the values of hyper-parameters, we use the grid-search method to conduct parameter sensitivity analysis of hyper-parameters $\alpha$, $\beta$, and $\gamma$ in three datasets. Finally, we illustrate the training curves of our method and use CAM to visualize the space-time regions that contribute to our predictions.

\subsubsection{Berkeley-MHAD \cite{ofli2013berkeley}}
This dataset consists of $11$ action classes performed by $12$ subjects with $5$ repetitions for each action. There are $12$ different camera views in total. The action data modalities include RGB videos, depth images, accelerators, and microphones. In this paper, we use RGB videos and accelerators. There are $7,900$ RGB videos and $6$ different accelerator modalities. Each accelerator modality has $658$ samples and the total number of accelerator samples is $3,948$. In all experiments, we use the first $7$ subjects for training and the last $5$ subjects for testing.

 \begin{table}[!t]
            \renewcommand{\arraystretch}{1}
             \caption{Implementation details for three benchmark datasets. Batch denotes the batch size, LR is the initial learning rate, DR is the decay ratio of the learning rate, DI is the decay iterations of the learning rate, Iters is the total iterations.}
              \label{Table 1}
            \centering
            \begin{tabular}{c|c|c|c|c|c|c}\hline
          Dataset&Modality&Batch&LR&DR&DI&Iters\\\hline
          \multirow{2}*{Berkeley-MHAD}&Teacher&8&0.0001&0.5&50&100\\
          &Student&8&0.001&0.1&20&30\\\hline
          \multirow{2}*{UTD-MHAD}&Teacher&16&0.0002&0.5&50&100\\
          &Student&16&0.001&0.5&50&100\\\hline
          \multirow{2}*{MMAct}&Teacher&16&0.0001&0.5&50&70\\
         &Student&32&0.001&0.5&30&60\\\hline
            \end{tabular}\vspace{-10pt}
 \end{table}

\subsubsection{UTD-MHAD \cite{chen2015utd}}
It consists of $27$ different actions performed by $8$ subjects with $4$ repetitions. This dataset has five modalities: RGB, depth, skeleton, Kinect, and inertial data. The vision-sensors data are captured by Kinect camera, while wearable-sensors data are captured by the inertial sensor. In this paper, we use RGB videos and two different wearable-sensors modalities (accelerator, gyroscope). Each modality has $861$ samples. Since each subject performs each action for $4$ times, we choose the first two samples of each action to form the training set and the remaining samples as the testing set.
\subsubsection{MMAct \cite{kong2019mmact}}

MMAct is a large-scale multi-modal action dataset consist of more than $36,000$ trimmed clips with seven modalities captured by $20$ subjects, which include RGB videos, keypoints, acceleration, gyroscope, orientation, Wi-Fi, and pressure signal. Each modality has $37$ action classes. This dataset is challenging as it contains $4$ camera views combining with random walks and occlusion scenes. In this paper, we use RGB videos and four different wearable-sensors modalities (accelerator-phone, accelerator-watch, gyroscope, and orientation). We use four different settings to evaluate this dataset: cross-subject, cross-view, cross-scene, and cross-session according to the train-test split strategy in \cite{kong2019mmact}.

For teacher networks, we used VGGNet16 \cite{simonyan2014very} as the backbone. For the student network,  we adopt multi-scale TRN \cite{zhou2018temporal} with BN-Inception pretrained on ImageNet because of its balance between accuracy and efficiency. In the multi-scale TRN, we set the dropout ratio as $0.8$ after the global pooling layer to reduce the effect of over-fitting, the number of segments is set as $8$ for Berkeley-MHAD and UTD-MHAD, while $3$ for the MMAct. The implementation details for Berkeley-MHAD, UTD-MHAD, and MMAct datasets are presented in Table \ref{Table 1}. All the experiments are conducted with two NVIDIA RTX 2080Ti GPUs using PyTorch \cite{paszke2019pytorch}. To synchronize two different modalities, we make the random seed for initializing these two dataloaders to be the same. For semantic representation extraction, we use Glove \cite{pennington2014glove} model and obtain $300$ dimensional semantic vectors for each action class names. We set hyper-parameters $\alpha$, $\beta$ and $\gamma$ in SAKDN according to the parameter sensitivity analysis in Section \uppercase\expandafter{\romannumeral4}.D.

\begin{table}[!t]
            \renewcommand{\arraystretch}{1}
             \caption{Performance comparison on Berkeley-MHAD.}
              \label{Table 2}
            \centering
            \begin{tabular}{c|c|c|c}\hline
          Type&Method&Modality&Accuracy\\\hline
          \multirow{3}*{VAR }&TSN \cite{wang2018temporal}&RGB videos&88.19\\
          &TRN \cite{zhou2018temporal}&RGB videos&95.32\\
          &TSM \cite{lin2019tsm}&RGB videos&96.87\\\hline
          \multirow{3}*{MMAR }&MKL \cite{ofli2013berkeley}&Accelerators+Depth&97.81\\
          &MPE \cite{shafaei2016real}&Accelerators+Depth&98.10\\
          &MOCAP \cite{ijjina2014human}&Accelerators+Depth&98.38\\\hline
          \multirow{7}*{KD }&Logits \cite{ba2014deep}&Accelerators+RGB Videos&97.93\\
          &Fitnet \cite{DBLP:journals/corr/RomeroBKCGB14}&Accelerators+RGB Videos&94.38\\
          &ST \cite{hinton2015distilling}&Accelerators+RGB Videos&95.99\\
          &AT \cite{DBLP:conf/iclr/ZagoruykoK17}&Accelerators+RGB Videos&97.99\\
          &RKD \cite{park2019relational}&Accelerators+RGB Videos&97.11\\
          &SP \cite{tung2019similarity}&Accelerators+RGB Videos&98.17\\
          &CC \cite{peng2019correlation}&Accelerators+RGB Videos&97.11\\\hline
          Proposed&SAKDN&Accelerators+RGB Videos&\textbf{99.33}\\\hline
            \end{tabular}
 \end{table}

 \begin{table}[!t]
            \renewcommand{\arraystretch}{1}
             \caption{Performance comparison on UTD-MHAD.}
              \label{Table 3}
            \centering
            \begin{tabular}{c|c|c|c}\hline
          Type&Method&Modality&Accuracy\\\hline
         \multirow{3}*{VAR }&TSN \cite{wang2018temporal}&RGB videos&92.54\\
          &TRN \cite{zhou2018temporal}&RGB videos&94.87\\
          &TSM \cite{lin2019tsm}&RGB videos&94.17\\\hline
          \multirow{3}*{MMAR }&CRC \cite{chen2015utd}&Acc+Gyro+Depth&79.10\\
          &CRC-2 \cite{chen2015real}&Acc+Gyro+Depth&97.20\\
          &CNN+LSTM \cite{dawar2018data}&Acc+Gyro+Depth&89.20\\
          &MFLF \cite{ehatisham2019robust}&Acc+Gyro+RGB Videos&98.20\\\hline
          \multirow{7}*{KD }&Logits \cite{ba2014deep}&Acc+Gyro+RGB Videos&97.20\\
          &Fitnet \cite{DBLP:journals/corr/RomeroBKCGB14}&Acc+Gyro+RGB Videos&90.20\\
          &ST \cite{hinton2015distilling}&Acc+Gyro+RGB Videos&97.90\\
          &AT \cite{DBLP:conf/iclr/ZagoruykoK17}&Acc+Gyro+RGB Videos&95.80\\
          &RKD \cite{park2019relational}&Acc+Gyro+RGB Videos&96.73\\
          &SP \cite{tung2019similarity}&Acc+Gyro+RGB Videos&94.40\\
          &CC \cite{peng2019correlation}&Acc+Gyro+RGB Videos&94.87\\\hline
          Proposed&SAKDN&Acc+Gyro+RGB Videos&\textbf{98.60}\\\hline
            \end{tabular}
 \end{table}

 \begin{table}[!t]
            \renewcommand{\arraystretch}{1}
             \caption{Performance comparison on MMAct.}
              \label{Table 4}
            \centering
            \begin{tabular}{c|c|c|c|c|c}\hline
            \multirow{2}*{Method}&\multirow{2}*{Modality}&Cross &Cross&Cross&Cross\\
             ~&~&subject&view&scene&session \\\hline
            TSN\cite{wang2018temporal}&RGB videos&59.50&54.37&51.21&68.65\\
           TRN \cite{zhou2018temporal}&RGB videos&66.56&65.51&60.03&71.95\\
          TSM \cite{lin2019tsm}&RGB videos&70.12&67.22&66.04&81.32\\\hline
          SMD  \cite{hinton2015distilling}&A+RGB&63.89&66.31&61.56&71.23\\
          MMD  \cite{kong2019mmact}&A+G+O+RGB&64.33&68.19&62.23&72.08\\
          MMAD  \cite{kong2019mmact}&A+G+O+RGB&66.45&70.33&64.12&74.58\\\hline
          Logits \cite{ba2014deep}&A+G+O+RGB&65.06&60.94&57.92&74.14\\\
          Fitnet \cite{DBLP:journals/corr/RomeroBKCGB14}&A+G+O+RGB&33.96&30.14&18.88&35.87\\\
          ST \cite{hinton2015distilling}&A+G+O+RGB&64.45&60.39&58.72&74.80\\\
          AT \cite{DBLP:conf/iclr/ZagoruykoK17}&A+G+O+RGB&65.59&60.30&55.92&74.28\\\
          RKD \cite{park2019relational}&A+G+O+RGB&65.54&61.67&55.38&75.05\\\
          SP \cite{tung2019similarity}&A+G+O+RGB&65.16&60.76&57.48&74.41\\\
          CC \cite{peng2019correlation}&A+G+O+RGB&65.60&59.59&59.65&73.98\\\hline
          SAKDN&A+G+O+RGB&\textbf{77.23}&\textbf{73.48}&\textbf{66.38}&\textbf{82.77}\\\hline
            \end{tabular}\vspace{-10pt}
 \end{table}

 \begin{table*}[!t]
            \renewcommand{\arraystretch}{0.9}\renewcommand\tabcolsep{9.0pt}
             \caption{Average accuracies (\%) on Berkeley-MHAD dataset. W/O denotes Without, A denotes Accelerator. The number in parenthesis means decreased accuracy over the proposed SAKDN.}
              \label{Table 5}
            \centering
            \begin{tabular}{l|c|c|c|c|c}\hline
            \multirow{2}*{Method}&Teacher&Student&Train&Test&\multirow{2}*{Accuracy}\\
           ~&Backbone&Backbone&Modality&Modality&\\\hline
          Teacher-Acc1 (SKDN)&VGG16&VGG16&Accelerator 1&Accelerator 1&78.18\\
          Teacher-Acc1 (AKDN)&VGG16&VGG16&Accelerator 1&Accelerator 1&74.90\\
          Teacher-Acc1 (SAKDN)&VGG16&VGG16&Accelerator 1&Accelerator 1&\textbf{81.09}\\\hline
          Teacher-Acc2 (SKDN)&VGG16&VGG16&Accelerator 2&Accelerator 2&75.63\\
          Teacher-Acc2 (AKDN)&VGG16&VGG16&Accelerator 2&Accelerator 2&73.45\\
          Teacher-Acc2 (SAKDN)&VGG16&VGG16&Accelerator 2&Accelerator 2&\textbf{82.90}\\\hline
          Teacher-Acc3 (SKDN)&VGG16&VGG16&Accelerator 3&Accelerator 3&71.63\\
          Teacher-Acc3 (AKDN)&VGG16&VGG16&Accelerator 3&Accelerator 3&68.72\\
          Teacher-Acc3 (SAKDN)&VGG16&VGG16&Accelerator 3&Accelerator 3&\textbf{75.27}\\\hline
          Teacher-Acc4 (SKDN)&VGG16&VGG16&Accelerator 4&Accelerator 4&76.00\\
          Teacher-Acc4 (AKDN)&VGG16&VGG16&Accelerator 4&Accelerator 4&70.54\\
          Teacher-Acc4 (SAKDN)&VGG16&VGG16&Accelerator 4&Accelerator 4&\textbf{80.36}\\\hline
          Teacher-Acc5 (SKDN)&VGG16&VGG16&Accelerator 5&Accelerator 5&52.72\\
          Teacher-Acc5 (AKDN)&VGG16&VGG16&Accelerator 5&Accelerator 5&51.27\\
          Teacher-Acc5 (SAKDN)&VGG16&VGG16&Accelerator 5&Accelerator 5&\textbf{55.27}\\\hline
          Teacher-Acc6 (SKDN)&VGG16&VGG16&Accelerator 6&Accelerator 6&50.90\\
          Teacher-Acc6 (AKDN)&VGG16&VGG16&Accelerator 6&Accelerator 6&46.18\\
          Teacher-Acc6 (SAKDN)&VGG16&VGG16&Accelerator 6&Accelerator 6&\textbf{54.54}\\\hline
          Multi-Teachers (SKDN)&VGG16&VGG16&A1+A2+A3+A4+A5+A6&A1+A2+A3+A4+A5+A6&89.09\\
          Multi-Teachers (AKDN)&VGG16&VGG16&A1+A2+A3+A4+A5+A6&A1+A2+A3+A4+A5+A6&90.54\\
          Multi-Teachers (SAKDN)&VGG16&VGG16&A1+A2+A3+A4+A5+A6&A1+A2+A3+A4+A5+A6&\textbf{92.00}\\\hline
          Student (Baseline)&BNInception&BNInception&RGB videos&RGB videos&$95.32\textbf{ (-4.01)}$\\\hline
          SKDN (W/O SPAMFM)&VGG16&BNInception&A1+A2+A3+A4+A5+A6+RGB&RGB videos&98.11\textbf{ (-1.22)}\\
           KDN (W/O ST)&VGG16&BNInception&A1+A2+A3+A4+A5+A6+RGB&RGB videos&98.14\textbf{ (-1.19)}\\
           SADN (W/O GSDM)&VGG16&BNInception&A1+A2+A3+A4+A5+A6+RGB&RGB videos&98.48\textbf{ (-0.85)}\\
            AKDN (W/O SP)&VGG16&BNInception&A1+A2+A3+A4+A5+A6+RGB&RGB videos&97.63\textbf{ (-1.70)}\\
           SAKDN&VGG16&BNInception&A1+A2+A3+A4+A5+A6+RGB&RGB videos&\textbf{99.33}\\\hline
            \end{tabular}\vspace{-10pt}
 \end{table*}

\subsection{Comparison with State-of-the-Art Methods}

We compare the performance of our SAKDN with state-of-the-art knowledge distillation (KD) methods \cite{ba2014deep,DBLP:journals/corr/RomeroBKCGB14,hinton2015distilling,DBLP:conf/iclr/ZagoruykoK17,park2019relational,tung2019similarity,peng2019correlation},
vision-based action recognition (VAR) methods \cite{wang2018temporal,zhou2018temporal,lin2019tsm}, and multi-modal action recognition (MMAR) methods \cite{ofli2013berkeley,shafaei2016real,ijjina2014human,chen2015utd,chen2015real,dawar2018data,ehatisham2019robust,kong2019mmact}.
The comparison results on three datasets are shown in Table \ref{Table 2}, Table \ref{Table 3} and Table \ref{Table 4}, respectively. For these VAR and KD methods, we use the shared codes, and the parameters are selected based on the default setting. Since \cite{kong2019mmact} is the only existing multi-modal action recognition method for the MMAct dataset which uses F-measure to evaluate the performance, we also adopt the F-measure in the MMAct dataset to make a fair comparison.

In Table \ref{Table 2} and Table \ref{Table 3}, the SAKDN achieves better performance than all the comparison action recognition methods, multi-modal action recognition methods, and knowledge distillation methods. This validates that our SAKDN can effectively improve vision-sensor based action recognition performance by integrating SPAMFM, GSDM, and SP into a unified end-to-end adaptive knowledge distillation framework. From Table \ref{Table 4}, we can see that the SAKDN performs better than most of the comparison action recognition methods, multi-modal action recognition methods, and knowledge distillation methods. To be noticed, the TSM \cite{lin2019tsm} achieves a comparable performance with the SAKDN. This is because the TSM shifts part of the channels along the temporal dimension and thus facilitates information exchanged among neighboring frames. Although the MMAD \cite{kong2019mmact} proposed a multi-modality distillation model to transfer the knowledge from wearable-sensors to vision-sensors, it only uses raw one-dimensional time-series sensor signals without considering virtual image generation of wearable-sensor data and the semantic relationship. When using the SAKDN, the performance is the best among all the comparison methods. This validates the effectiveness of our adaptive knowledge distillation framework for sensor-to-vision heterogenous action recognition by integrating Gramian Angular Field (GAF) based virtual image generation, SPAMFM, GSDM into a unified end-to-end deep learning framework.

\subsection{Ablation Study}
To evaluate the contribution of the SPAMFM, the ST, the GSDM, and the SP, we construct four different algorithms based on the SAKDN.
\textbf{Student (Baseline)}: our student network trained with only RGB videos. \textbf{Multi-Teachers}: our teacher networks trained with all wearable-sensor modalities.
\textbf{SKDN}: our SAKDN without Similarity-Preserving Adaptive Multi-modal Fusion Module (SPAMFM).  \textbf{KDN}:
our SAKDN without soft-target loss. \textbf{SADN}: our SAKDN without Graph-guided Semantically Discriminative Mapping (GSDM). \textbf{AKDN}: our SAKDN without Semantic Preserving (SP) for
both teacher and student networks. \textbf{SAKDN}: our proposed Semantics-aware Adaptive Knowledge Distillation Networks.

\begin{table*}[!t]
            \renewcommand{\arraystretch}{0.9}\renewcommand\tabcolsep{17.0pt}
             \caption{Average accuracies (\%) on UTD-MHAD dataset. W/O denotes Without. The number in parenthesis means decreased accuracy over the proposed SAKDN.}
              \label{Table 6}
            \centering
            \begin{tabular}{l|c|c|c|c|c}\hline
           \multirow{2}*{Method}&Teacher&Student&Train&Test&\multirow{2}*{Accuracy}\\
           ~&Backbone&Backbone&Modality&Modality&\\\hline
          Teacher-Acc (SKDN)&VGG16&VGG16&Accelerator&Accelerator&96.27\\
          Teacher-Acc (AKDN)&VGG16&VGG16&Accelerator&Accelerator&91.84\\
          Teacher-Acc (SAKDN)&VGG16&VGG16&Accelerator&Accelerator&\textbf{97.66}\\\hline
           Teacher-Gyo (SKDN)&VGG16&VGG16&Gyroscope&Gyroscope&93.93\\
           Teacher-Gyo (AKDN)&VGG16&VGG16&Gyroscope&Gyroscope&92.77\\
          Teacher-Gyo (SAKDN)&VGG16&VGG16&Gyroscope&Gyroscope&\textbf{94.87}\\\hline
           Multi-Teachers (SKDN)&VGG16&VGG16&Acc+Gyo&Acc+Gyo&96.27\\
            Multi-Teachers (AKDN)&VGG16&VGG16&Acc+Gyo&Acc+Gyo&97.43\\
           Multi-Teachers (SAKDN)&VGG16&VGG16&Acc+Gyo&Acc+Gyo&\textbf{98.83}\\\hline
          Student (Baseline)&BNInception&BNInception&RGB videos&RGB videos&94.87\textbf{ (-3.73)}\\\hline
          SKDN (W/O SPAMFM)&VGG16&BNInception&Acc+Gyo+RGB&RGB videos&97.43\textbf{ (-1.17)}\\
          KDN  (W/O ST)&VGG16&BNInception&Acc+Gyo+RGB&RGB videos&96.27\textbf{ (-2.33)}\\
           SADN (W/O GSDM)&VGG16&BNInception&Acc+Gyo+RGB&RGB videos&97.66\textbf{ (-0.94)}\\
           AKDN (W/O SP)&VGG16&BNInception&Acc+Gyo+RGB&RGB videos&96.96\textbf{ (-1.64)}\\
           SAKDN&VGG16&BNInception&Acc+Gyo+RGB&RGB videos&\textbf{98.60}\\\hline
            \end{tabular}
 \end{table*}
 \begin{table*}[!t]
            \renewcommand{\arraystretch}{0.9}\renewcommand\tabcolsep{6.0pt}
             \caption{Average accuracies (\%) on MMAct dataset. W/O denotes Without. The number in parenthesis means decreased accuracy over the proposed SAKDN.}
              \label{Table 7}
            \centering
            \begin{tabular}{l|c|c|c|c|c|c|c|c}\hline
           \multirow{2}*{Method}&Teacher&Student&Train&Test&Cross&Cross &Cross &Cross \\
           ~&Backbone&Backbone&Modality&Modality& Subject&View&Scene&Session\\\hline
          Teacher-Ap (SKDN)&VGG16&VGG16&Acc-phone&Acc-phone&49.54&56.65&55.44&56.81\\
          Teacher-Ap (AKDN)&VGG16&VGG16&Acc-phone&Acc-phone&43.41&52.30&49.22&53.57\\
          Teacher-Ap (SAKDN)&VGG16&VGG16&Acc-phone&Acc-phone&\textbf{52.34}&\textbf{59.82}&\textbf{57.15}&\textbf{59.38}\\\hline
          Teacher-Aw (SKDN)&VGG16&VGG16&Acc-watch&Acc-watch&44.23&49.97&63.26&16.50\\
          Teacher-Aw (AKDN)&VGG16&VGG16&Acc-watch&Acc-watch&37.08&47.08&60.54&16.44\\
          Teacher-Aw (SAKDN)&VGG16&VGG16&Acc-watch&Acc-watch&\textbf{44.83}&\textbf{53.14}&\textbf{69.42}&\textbf{18.58}\\\hline
          Teacher-Gyo (SKDN)&VGG16&VGG16&Gyroscope&Gyroscope&44.70&37.83&50.40&56.14\\
          Teacher-Gyo (AKDN)&VGG16&VGG16&Gyroscope&Gyroscope&41.52&37.74&47.85&51.39\\
          Teacher-Gyo (SAKDN)&VGG16&VGG16&Gyroscope&Gyroscope&\textbf{52.98}&\textbf{40.86}&\textbf{56.52}&\textbf{59.66}\\\hline
          Teacher-Ori (SKDN)&VGG16&VGG16&Orientation&Orientation&42.87&55.09&53.78&57.70\\
          Teacher-Ori (AKDN)&VGG16&VGG16&Orientation&Orientation&40.72&54.20&51.29&53.74\\
          Teacher-Ori (SAKDN)&VGG16&VGG16&Orientation&Orientation&\textbf{47.12}&\textbf{60.60}&\textbf{58.71}&\textbf{61.56}\\\hline
          Multi-Teachers (SKDN)&VGG16&VGG16&Ap+Aw+Gyo+Ori&Ap+Aw+Gyo+Ori&67.45&65.66&78.72&68.77\\
          Multi-Teachers (AKDN)&VGG16&VGG16&Ap+Aw+Gyo+Ori&Ap+Aw+Gyo+Ori&66.64&65.88&79.24&66.53\\
          Multi-Teachers (SAKDN)&VGG16&VGG16&Ap+Aw+Gyo+Ori&Ap+Aw+Gyo+Ori&\textbf{68.69}&\textbf{68.22}&\textbf{81.61}&\textbf{70.11}\\\hline
           \multirow{2}*{Student (Baseline)}&\multirow{2}*{BNInception}&\multirow{2}*{BNInception}&\multirow{2}*{RGB videos}&\multirow{2}*{RGB videos}&68.41&65.25&56.33&76.79\\
           &&&&&\textbf{(-2.70)}&\textbf{(-3.33)}&\textbf{(-7.08)}&\textbf{(-4.98)}\\\hline
          \multirow{2}*{SKDN (W/O SPAMFM)}&\multirow{2}*{VGG16}&\multirow{2}*{BNInception}&\multirow{2}*{Ap+Aw+Gyo+Ori+RGB}&\multirow{2}*{RGB videos}&70.38&67.42&57.69&76.96\\
            &&&&&\textbf{(-0.73)}&\textbf{(-1.16)}&\textbf{(-5.72)}&\textbf{(-4.81)}\\\hline
              \multirow{2}*{KDN  (W/O ST)}& \multirow{2}*{VGG16}& \multirow{2}*{BNInception}&\multirow{2}*{Ap+Aw+Gyo+Ori+RGB}&\multirow{2}*{RGB videos}&70.45&67.26&61.90&80.79\\
                         &&&&&\textbf{(-0.66)}&\textbf{(-1.32)}&\textbf{(-1.51)}&\textbf{(-0.98)}\\\hline
           \multirow{2}*{SADN (W/O GSDM)}&\multirow{2}*{VGG16}&\multirow{2}*{BNInception}&\multirow{2}*{Ap+Aw+Gyo+Ori+RGB}&\multirow{2}*{RGB videos}&69.11&65.16&56.86&79.53\\
            &&&&&\textbf{(-2.00)}&\textbf{(-3.42)}&\textbf{(-6.55)}&\textbf{(-2.24)}\\\hline
           \multirow{2}*{AKDN (W/O SP)}&\multirow{2}*{VGG16}&\multirow{2}*{BNInception}&\multirow{2}*{Ap+Aw+Gyo+Ori+RGB}&\multirow{2}*{RGB videos}&70.63&64.03&62.48&79.63\\
            &&&&&\textbf{(-0.48)}&\textbf{(-4.55)}&\textbf{(-0.93)}&\textbf{(-2.14)}\\\hline
            SAKDN& VGG16&BNInception&Ap+Aw+Gyo+Ori+RGB&RGB videos&\textbf{71.11}&\textbf{68.58}&\textbf{63.41}&\textbf{81.77}\\\hline
            \end{tabular}\vspace{-10pt}
 \end{table*}

The average accuracies on Berkeley-MHAD, UTD-MHAD, and MMAct datasets are shown in Table \ref{Table 5}, \ref{Table 6} and \ref{Table 7}, respectively. In Table \ref{Table 5}, the SAKDN for multi-teachers achieves better performance than that of the single teacher modality (Acc1-Acc6). Additionally, the SAKDN for six teacher modalities achieve better performance than that of the SKDN, these validate that our proposed SPAMFM can effectively fuse the complementary knowledge among different wearable-sensors. Without soft-target loss, the KDN performs worse than the SAKDN. This shows the importance of the ST in conducting knowledge transfer at the last fully-connected layers between teacher and student networks. In addition, the SAKDN also performs better than the AKDN, this verifies that the semantics-preserving part indeed acts as an effective guide for knowledge transfer. The student model with only RGB input (Baseline) achieves better performance than that of the teacher models because the wearable-sensors on Berkeley-MHAD dataset lack color and texture information which may degrade their representative abilities. Introducing different accelerators to the student model improves the performance from $95.32\%$ to $99.33\%$, which validates the existence of complementary knowledge between
wearable-sensors and vision-sensor modalities. In SAKDN, a more significant improvement of performance is achieved than SKDN, KDN, SADN, and AKDN in video action recognition. In addition, the performance of SKDN, KDN, SADN, AKDN are all better than that of the student baseline, which verifies that the SPAMFM, ST, GSDM, and SP are complementary and essential.

From Table \ref{Table 6}, we can see that the teacher-acc, teacher-gyo and multi-teachers using SAKDN outperform the SKDN and AKDN. This validates that the SPAMFM can fully utilize
complementary knowledge from multiple teacher modalities and the SP can guide the knowledge transfer using the similar semantic relationship between teacher and student modalities. Moreover, the performance
of the SKDN, KDN, SADN, and AKDN are all better than that of the student baseline, which validates the effectiveness of SPAMFM, ST, GSDM, and SP. Among SKDN, KDN, SADN, AKDN, and SAKDN, the SAKDN achieves the best performance and improves the video action recognition performance from $94.87\%$ (baseline) to $98.60\%$. This validates that our SAKDN can effectively transfer the knowledge from wearable-sensor modalities to vision-sensor modalities by integrating four complementary modules, SPAMFM, ST, GSDM, and SP.

To make a fair comparison with other methods in MMAct dataset, we use four different settings. 1) cross-subject: samples from $80\%$ of the subjects are used for training and the remaining $20\%$ for testing; 2) cross-view: samples from $3$ views are used for training and the remaining one for testing; 3) cross-scene: samples from the scenes without occlusion are used for training and the samples with occlusion scene for testing; 4) cross-session: samples from top $80\%$ sessions in ascending order for each subject are used for training and the remaining sessions for testing. The results for different settings on the MMAct dataset are shown in Table \ref{Table 7}. In cross-view and cross-scene settings, the multi-teachers achieves better performance than that of the baseline. This is because wearable-sensor based action data are more robust to the occlusion and appearance variations caused by the camera viewpoint and scene change. Since appearance and texture information are important for action recognition in cross-subject and cross-session settings, the baseline model fully utilizes appearance information and performs better than the multi-teachers which lack texture and color information to discriminate different human subjects. Base on these observations, we can validate that wearable-sensors and vision-sensors modalities are related and complementary. To be noticed, the performance drops significantly compared with that of the SAKDN when using SADN (without GSDM), this validates that the GSDM is significantly important for knowledge transfer from wearable-sensors to vision-sensors. In the cross-view setting where the appearance of human actions varies a lot because of the changing camera views, the performance of SADN (without GSDM) and the AKDN (without SP) both perform worse than the baseline. While in SAKDN, the performance increase by $3.33\%$. This shows that the GSDM and SP are both important for addressing the cross-view challenge. More importantly, our SAKDN outperforms both the multi-teacher model and the baseline model under all settings, which verifies that the SAKDN can effectively exploit complementary knowledge between wearable-sensors and vision-sensors action data and then improve the video action recognition performance in the wild. Among all the settings, the SAKDN performs significantly better than that of the  SKDN, KDN, SADN, and AKDN, which demonstrates that the SPAMFM, ST, GSDM, and SP are all complementary and essential.

\begin{figure}[!t]
\centering
\includegraphics[scale=0.5]{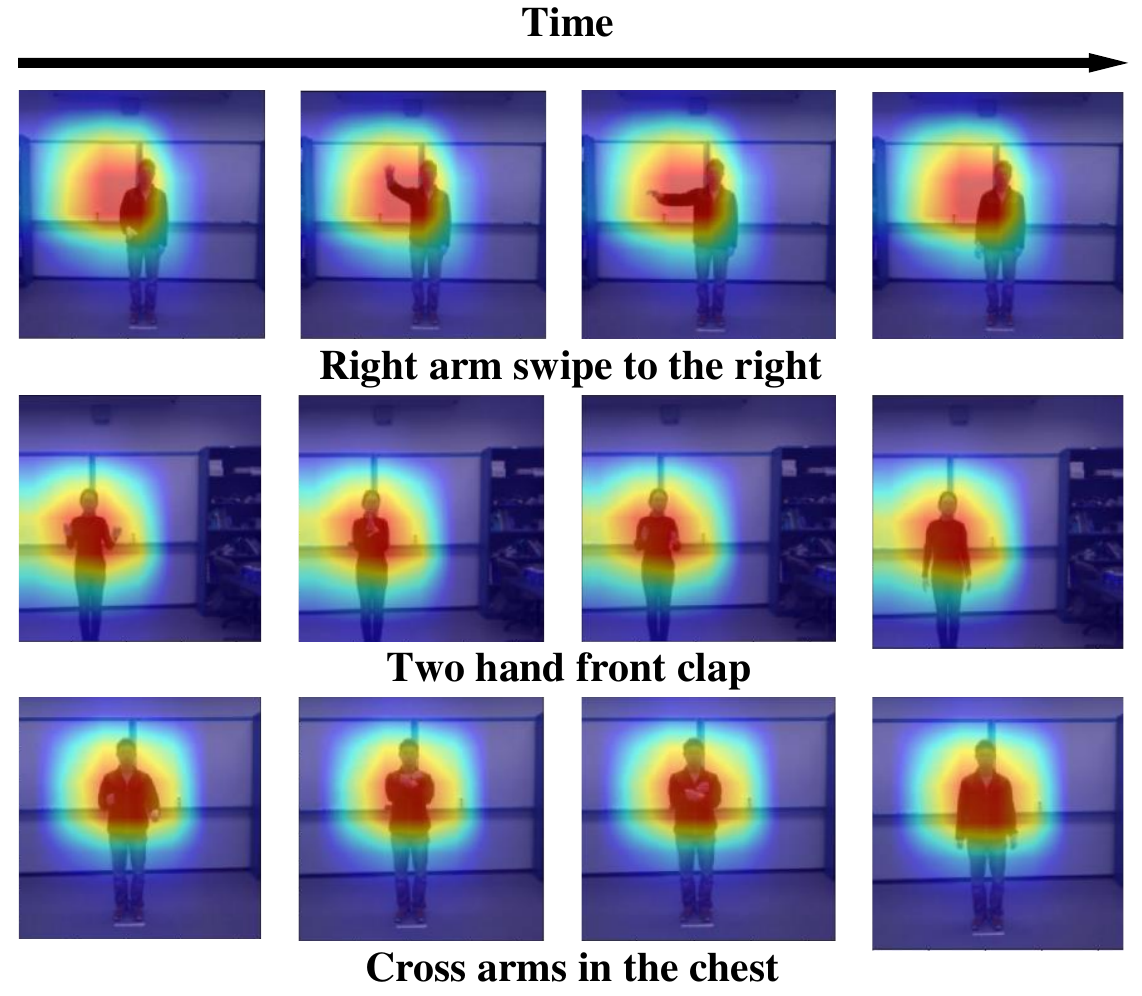}
\caption{Visualization of GSDM in UTD-MHAD dataset. The maps highlight the discriminative region for action recognition.  }
\vspace{-10pt}
\label{Fig6}
\end{figure}

To get an intuitive understanding of the effectiveness of ``ablation" video frames and what Graph-guided Semantically Discriminative Mapping (GSDM) learns, the GSDM generated by the student network (video modality) in the UTD-MHAD dataset is visualized in Fig. \ref{Fig6}. To facilitate visualization, the GSDM of the layer inception5b is resized to the input size and then used to weight each channel of the corresponding input video frames. These examples in Fig. \ref{Fig6} indicate that the GSDM can effectively highlight the important regions for predicting the semantic concept, which can be considered as an effective intermediate visual pattern for knowledge distillation.

\begin{figure*}[!t]
\centering
\includegraphics[scale=0.35]{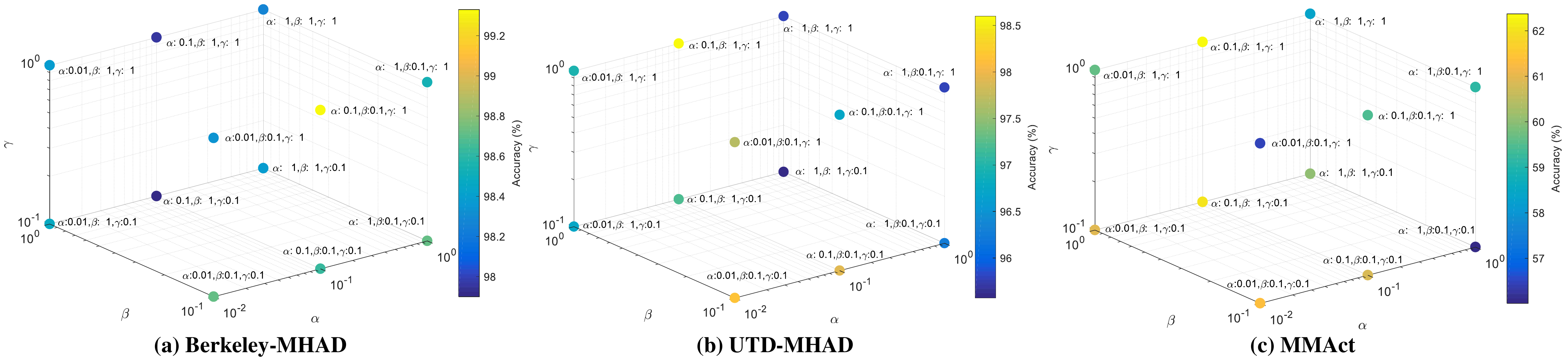}
\caption{ Four-dimensional scatter diagram for parameter sensitivity analysis of $\alpha$, $\beta$ and $\gamma$. }
\label{Fig7}
\end{figure*}

\begin{figure*}[!t]
\centering
\includegraphics[scale=0.28]{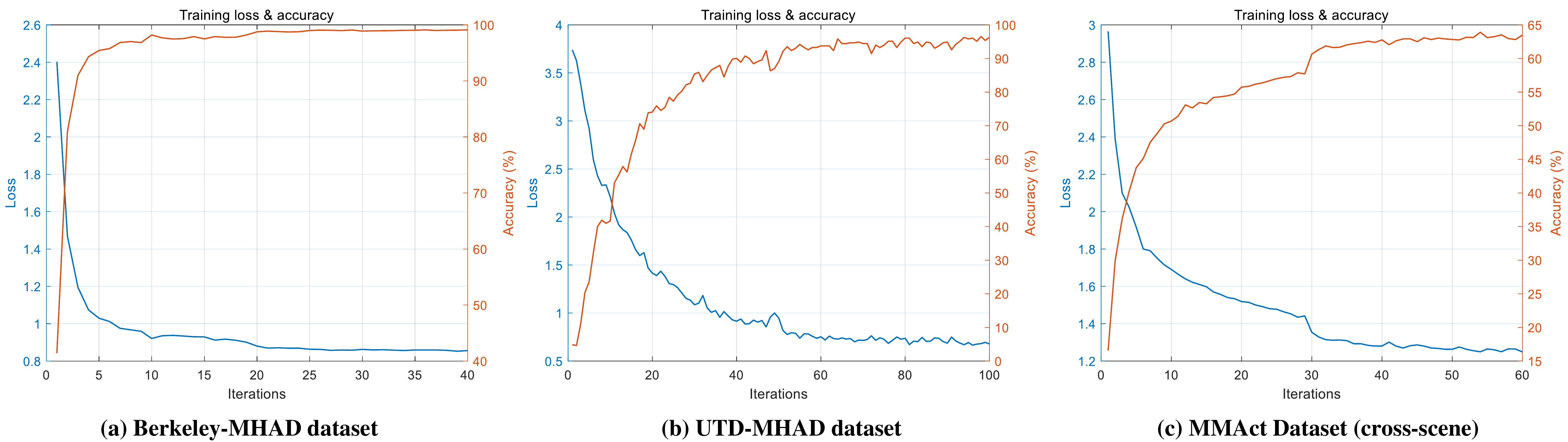}
\caption{Training loss and accuracy curves for Berkeley-MHAD, UTD-MHAD and MMAct datasets.  }
\vspace{-10pt}
\label{Fig8}
\end{figure*}

\subsection{Effect of Different Transfer Layers and Backbones}

To validate whether our SAKDN could generalize to different selected layers between teacher and student networks,
we evaluate the performance of the SAKDN in the UTD-MHAD dataset using different combinations of $\mathcal{L}_{distill}^T$ and $\mathcal{L}_{distill}^S$
in Eq. (\ref{eq21}) and (\ref{eq22}). In addition, we use different backbones (BNInception, ResNet18, and ResNet50) for the student network to measure
the generalization ability of SAKDN in different student backbones, as shown in Table \ref{Table 8}. For BNIneption, the performances of using different transfer layers are all better than
that of the baseline ($94.87\%$). This validates that our SAKDN can conduct knowledge distillation across different layers between teacher and student networks. To be noticed, for the same backbone, the performance gap of using different numbers of transfer layers is marginal. This shows that our SAKDN can generalize well to different levels of transfer layers.
And the performance is the best when we use $\mathcal{L}_{distill}^T=\{\textrm{c}_1^2, \textrm{c}_2^2, \textrm{c}_3^3, \textrm{c}_4^3,  \textrm{c}_5^3\}$ and $\mathcal{L}_{distill}^S=\{\textrm{c}2, \textrm{I}{3c}, \textrm{I}{4c},  \textrm{I}{5a}, \textrm{I}{5b}\}$. When using different backbones, ResNet18, ResNet50 and BNInception all perform better than the baseline method, which verifies the generalization ability of the SAKDN in different backbones. Among the ResNet18, ResNet50 and BNInception, both the ResNet50 and BNInception achieve better performance than that of the ResNet18, which attributes to the simpler architecture of  the ResNet18 for representation learning.

\begin{table}[!t]
            \renewcommand{\arraystretch}{1.2}
             \caption{Average accuracies (\%) on UTD-MHAD dataset for different transfer layers and student backbones, where c denotes convolutional layer, I denotes inception layer.}
              \label{Table 8}
            \centering
            \begin{tabular}{c|l|l|c}\hline
            Student&Teacher &Student&\multirow{2}*{Acc}\\
             Backbone&Layers&Layers&~ \\\hline
           BNInception&$\{ \textrm{c}_5^3\}$&$\{  \textrm{I}{5a}\}$&97.66\\
           BNInception&$\{ \textrm{c}_5^3\}$&$\{ \textrm{I}{5b}\}$&97.90\\
           BNInception&$\{ \textrm{c}_4^3,  \textrm{c}_5^3\}$&$\{ \textrm{I}{4d}, \textrm{I}{5a}\}$&97.66\\
           BNInception&$\{\textrm{c}_4^3,  \textrm{c}_5^3\}$&$\{ \textrm{I}{5a}, \textrm{I}{5b}\}$&97.43\\
           BNInception&$\{ \textrm{c}_3^3, \textrm{c}_4^3,  \textrm{c}_5^3\}$&$\{ \textrm{I}{4b},  \textrm{I}{4d}, \textrm{I}{5a}\}$&96.27\\
           BNInception&$\{ \textrm{c}_3^3, \textrm{c}_4^3,  \textrm{c}_5^3\}$&$\{ \textrm{I}{4c},  \textrm{I}{5a}, \textrm{I}{5b}\}$&96.27\\
           BNInception&$\{\textrm{c}_2^2, \textrm{c}_3^3, \textrm{c}_4^3,  \textrm{c}_5^3\}$&$\{\textrm{I}{3c}, \textrm{I}{4b},  \textrm{I}{4d}, \textrm{I}{5a}\}$&98.36\\
           BNInception&$\{\textrm{c}_2^2, \textrm{c}_3^3, \textrm{c}_4^3,  \textrm{c}_5^3\}$&$\{\textrm{I}{3c}, \textrm{I}{4c},  \textrm{I}{5a}, \textrm{I}{5b}\}$&97.90\\\hline
           ResNet18&$\{\textrm{c}_1^2, \textrm{c}_2^2, \textrm{c}_3^3, \textrm{c}_4^3,  \textrm{c}_5^3\}$&$\{\textrm{c}_1^1, \textrm{c}_2^1, \textrm{c}_3^1,  \textrm{c}_4^1, \textrm{c}_5^1\}$&96.73\\
           ResNet50&$\{\textrm{c}_1^2, \textrm{c}_2^2, \textrm{c}_3^3, \textrm{c}_4^3,  \textrm{c}_5^3\}$&$\{\textrm{c}_1^1, \textrm{c}_2^1, \textrm{c}_3^{1},  \textrm{c}_4^{1}, \textrm{c}_5^1\}$&98.36\\
           BNInception&$\{\textrm{c}_1^2, \textrm{c}_2^2, \textrm{c}_3^3, \textrm{c}_4^3,  \textrm{c}_5^3\}$&$\{\textrm{I}{3a}, \textrm{I}{3c}, \textrm{I}{4b},  \textrm{I}{4d}, \textrm{I}{5a}\}$&97.20\\\hline
           ResNet18&$\{\textrm{c}_1^2, \textrm{c}_2^2, \textrm{c}_3^3, \textrm{c}_4^3,  \textrm{c}_5^3\}$&$\{\textrm{c}_1^1, \textrm{c}_2^4, \textrm{c}_3^4,  \textrm{c}_4^4, \textrm{c}_5^4\}$&95.33\\
           ResNet50&$\{\textrm{c}_1^2, \textrm{c}_2^2, \textrm{c}_3^3, \textrm{c}_4^3,  \textrm{c}_5^3\}$&$\{\textrm{c}_1^1, \textrm{c}_2^9, \textrm{c}_3^{12},  \textrm{c}_4^{18}, \textrm{c}_5^9\}$&97.90\\
           BNInception&$\{\textrm{c}_1^2, \textrm{c}_2^2, \textrm{c}_3^3, \textrm{c}_4^3,  \textrm{c}_5^3\}$&$\{\textrm{c}2, \textrm{I}{3c}, \textrm{I}{4c},  \textrm{I}{5a}, \textrm{I}{5b}\}$&\textbf{98.60}\\\hline
            \end{tabular}
 \end{table}
 
\begin{table}[!t]
            \renewcommand{\arraystretch}{1}\renewcommand\tabcolsep{9.0pt}
             \caption{Parameters sensitivity analysis of $\alpha$, $\beta$, $\gamma$ on Berkeley-MHAD, UTD-MHAD and MMAct datasets.}
              \label{Table 9}
            \centering
            \begin{tabular}{c|c|c|c|c|c}\hline
            \multirow{2}*{$\alpha$}&\multirow{2}*{$\beta$}&\multirow{2}*{$\gamma$}&Berkeley-&UTD-&MMAct \\
            &&&MHAD&MHAD&(cross-scene)\\\hline
           1&1&1&98.27&95.80&58.32\\\hline
           1&1&0.1&98.39&95.57&60.01\\\hline
           1&0.1&1&98.54&95.80&59.11\\\hline
           1&0.1&0.1&98.72&96.27&56.01\\\hline
           0.1&1&1&97.96&\textbf{98.60}&\textbf{62.37}\\\hline
           0.1&1&0.1&97.90&97.20&62.01\\\hline
           0.1&0.1&1&\textbf{99.33}&96.73&59.47\\\hline
           0.1&0.1&0.1&98.63&97.90&60.79\\\hline
           0.01&1&1&98.39&96.96&59.63\\\hline
           0.01&1&0.1&98.48&96.73&60.92\\\hline
           0.01&0.1&1&98.33&97.66&56.42\\\hline
           0.01&0.1&0.1&98.72&98.13&61.32\\\hline
            \end{tabular}\vspace{-10pt}
 \end{table}

\subsection{Parameter Sensitivity Analysis}
There are three hyper-parameters $\alpha$, $\beta$ and $\gamma$ in Eq. (\ref{eq25}). To learn how they influence the performance, we conduct the parameter sensitivity analysis on Berkeley-MHAD, UTD-MHAD, and MMAct datasets (cross-scene) using the grid-search method.
The parameter $\alpha$ get its values in $\{0.01,0.1,1\}$, and parameters $\beta$ and $\gamma$ get their values in $\{0.1,1\}$. Table \ref{Table 9} shows the performance of our SAKDN by allocating different values to the parameters $\alpha$, $\beta$, and $\gamma$. To have a more intuitive understanding of how we choose the optimal values for these parameters, we transform the table representation of parameter sensitivity in Table \ref{Table 9} to a four-dimensional scatter diagram with the polar coordinate, shown in Fig. \ref{Fig7}. From Table \ref{Table 9} and Fig. \ref{Fig7}, we can see that the optimal values for Berkeley-MHAD dataset are $\{0.1,0.1,1\}$, while for both UTD-MHAD and MMAct datasets, the optimal values are $\{0.1,1,1\}$. This means that the semantics-preserving should be attached more importance than the soft-target and GSDM, because the semantic relationship contributes a lot to multi-modal feature fusion, knowledge transfer, and representation learning for our SAKDN. For parameter $\alpha$, its optimal values are $0.1$ for all datasets. This is because the soft-target loss only conducts knowledge distillation at the last fully-connected layers, while for the GSDM and SP loss, they contribute to the knowledge transfer throughout all layers. For parameter $\beta$, the optimal value for the Berkeley-MHAD dataset is $0.1$, while for both UTD-MHAD and MMAct datasets, the optimal values are $1$. Since the Berkeley-MHAD dataset has six teacher modalities, which is larger than that of the  UTD-MHAD and MMAct datasets. Therefore, $\beta$ should be set to a moderate value to avoid overfitting to any one of the teacher modalities when conducting knowledge distillation. The value of $\gamma$ is set to $1$ for all datasets.

\begin{figure*}[!t]
\centering
\includegraphics[scale=0.36]{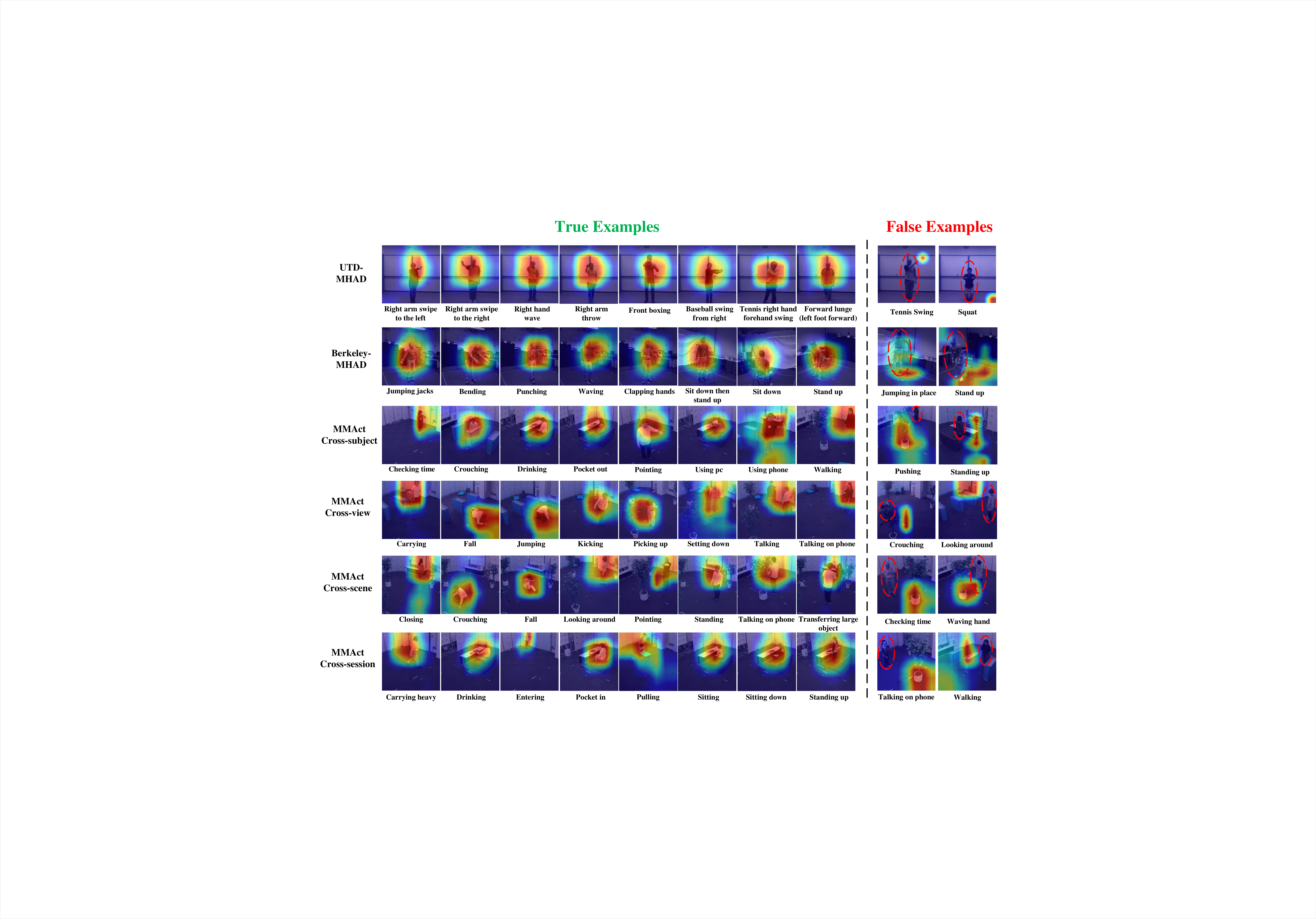}
\caption{CAM visualization examples of space-time regions for Berkeley-MHAD, UTD-MHAD, and MMAct datasets. We visualize such regions as overlaid heat-maps, where red and blue correspond to high and low activated regions, respectively. Interestingly, our model is able to pick up the salient mover in the presence of significant camera motion, background clutter, occlusion, and appearance variation. The red dashed ellipses in the false examples images denote the true space-time regions.}
\vspace{-10pt}
\label{Fig9}
\end{figure*}

\subsection{Visualization Analysis}

To show the training curves of the proposed method, we plot the training loss and accuracy for Berkeley-MHAD, UTD-MHAD, and MMAct datasets, as shown in Fig. \ref{Fig8}. The training loss steadily decreases and converges to a smaller value with the training iteration increasing. At the same time, the training accuracy increases with the number of training iterations increasing. This validates that our proposed loss function is effective with good convergence.

To gain a better understanding about which space-time regions contribute to our predictions, we follow the Class Activation Map (CAM) technique \cite{zhou2016learning} to visualize the energy of the last convolutional layer in the student network, before the global max and average pooling. We show some true examples and fail examples on Berkeley-MHAD, UTD-MHAD, and MMAct datasets to further analyze our proposed method. Fig. \ref{Fig9} depicts computed heat maps superimposed over sample video frames. In most cases, these examples portray a strong correlation between highly activated regions and the dominant movement in the scene, even when performing complex interactive actions in the presence of significant camera motion, background clutter, occlusion, and appearance variation. For example, in the fourth row, our model can capture dominant moving regions under different camera views. In the fifth row, with significant background clutter, occlusion, and appearance variation, our model can still concentrate on the body motion. These visualization results validate that our proposed model takes full advantage of both the vision-sensor modality and wearable-sensor modality in addressing the problem of background clutter, occlusion, and appearance variation, and thus is able to focus on the salient movement under these challenges. To be noticed, in some extremely challenging cases, our model fails to capture true moving regions in the video frames. For example, in the false examples, our model mistakes some background objects (pot plant, heavy luggage) as the body subjects due to the high appearance similarity, severe background clutter, and occlusion.

\section{Conclusion}
In this paper, we propose an end-to-end knowledge distillation framework, named Semantics-aware Adaptive Knowledge Distillation Networks (SAKDN), to adaptively distill the complementary knowledge from multiple wearable-sensors (teachers) to the vision-sensor (student), and concurrently improve the action recognition performance in vision-sensor modality (videos). To fully utilize the complementary knowledge from multiple teachers, we propose a novel plug-and-play module, named Similarity-Preserving Adaptive Multi-modal Fusion Module (SPAMFM), which integrates intra-modality similarity, semantic embeddings, and multiple relational knowledge to learn the global context representation and recalibrate the channel-wise features adaptively in each teacher network. To effectively exploit and transfer the knowledge of multiple well-trained teachers to the student, we propose a novel knowledge distillation module, named Graph-guided Semantically Discriminative Mapping (GSDM), which utilizes graph-guided ablation analysis to produce a visual explanation highlighting the important regions for predicting the semantic concept, and concurrently preserving respective interrelations of data. Extensive experiments on three benchmarks demonstrate the effectiveness of our SAKDN for adaptive knowledge transfer from wearable-sensors to vision-sensors.

\bibliographystyle{IEEEtran}
\bibliography{IEEEabrv,bibfile}

\begin{IEEEbiography}[{\includegraphics[width=1in,height=1.25in,clip,keepaspectratio]{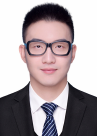}}]{Yang Liu}(M'21) is currently a Postdoctoral Researcher in the School of Computer Science and Engineering, Sun Yat-Sen University, working with Prof. Liang Lin. He received the Ph.D. degree in telecommunications and information systems from Xidian University, Xi'an, China, in June 2019, advised by Prof. Zhaoyang Lu. Before that, he received the B.S. degree in telecommunications engineering from Chang'an University, Xi'an, China, in 2014. His current research interests include video understanding, transfer learning and computer vision. He has been serving as a reviewer for numerous academic journals, including TNNLS, TMM, TCyb, TCSVT, THMS, SPL and PR. More information can be found on his personal website https://yangliu9208.github.io/home.
\end{IEEEbiography}

\begin{IEEEbiography}[{\includegraphics[width=1in,height=1.25in,clip,keepaspectratio]{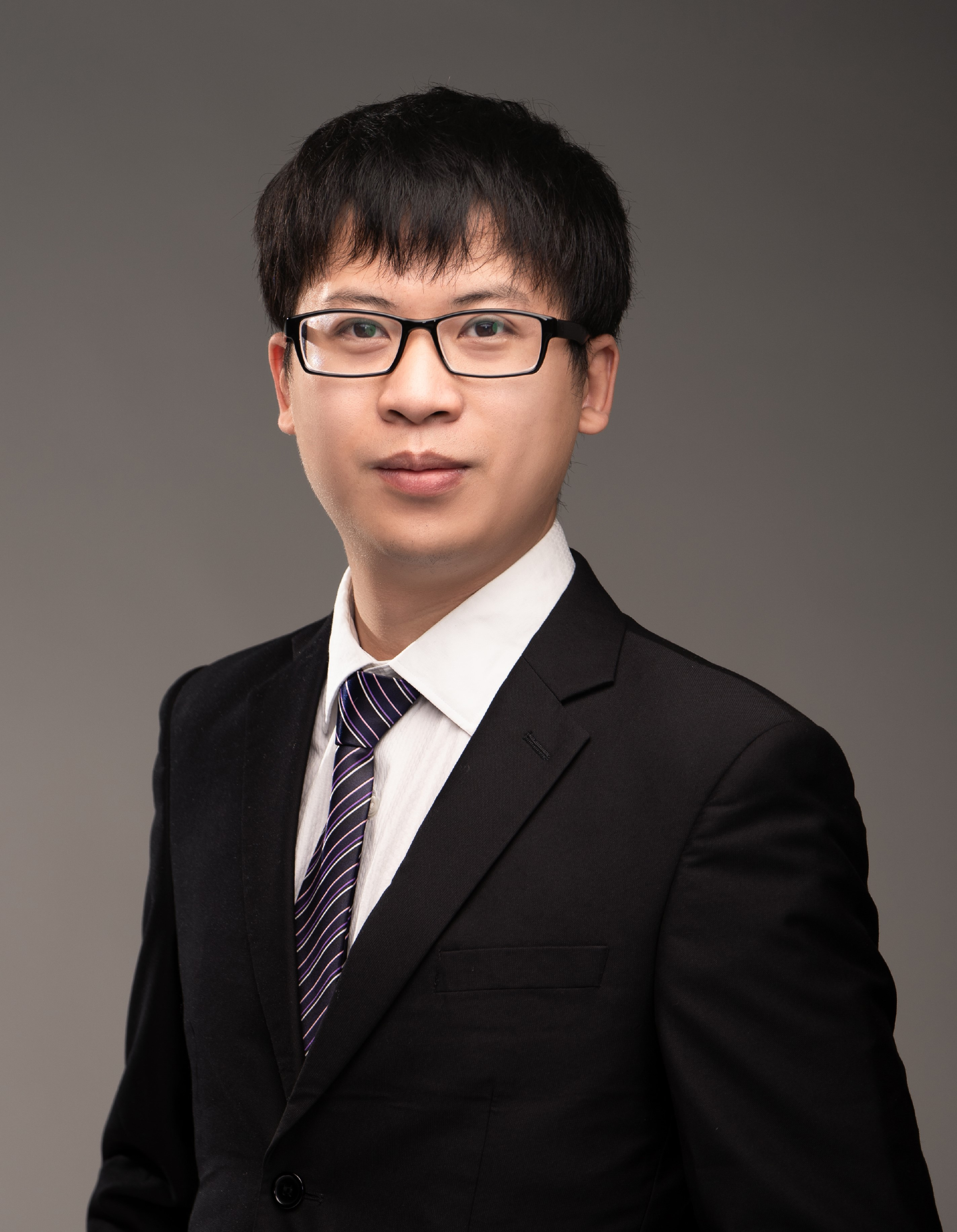}}]{Keze Wang }
received his B.S. degree in software engineering from Sun Yat-Sen University, Guangzhou, China, in 2012. He obtained my Ph.D. degree with honors from the School of Data and Computer Science at Sun Yat-Sen University in December 2017, advised by Prof. Liang Lin. He obtained dual PhD awards in the Department of Computing of the Hong Kong Polytechnic University in March 2019, advised by Prof. Lei Zhang. His current research interests include computer vision and machine learning. More information can be found on his personal website https://kezewang.com.
\end{IEEEbiography}

\begin{IEEEbiography}[{\includegraphics[width=1in,height=1.25in,clip,keepaspectratio]{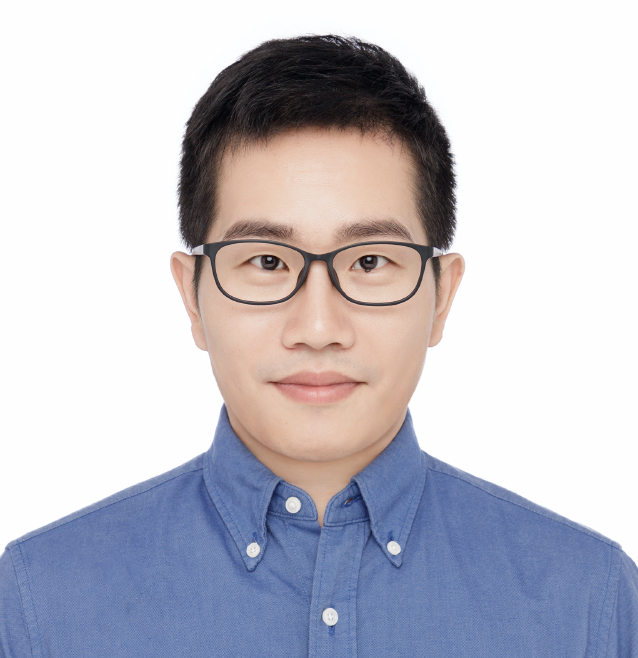}}]{Guanbin Li}(M'15) is currently an associate professor in School of Computer Science and Engineering, Sun Yat-Sen University. He received his PhD degree from the University of Hong Kong in 2016. His current research interests include computer vision, image processing, and deep learning. He is a recipient of ICCV 2019 Best Paper Nomination Award. He has authorized and co-authorized on more than 70 papers in top-tier academic journals and conferences. He serves as an area chair for the conference of VISAPP. He has been serving as a reviewer for numerous academic journals and conferences such as TPAMI, IJCV, TIP, TMM, TCyb, CVPR, ICCV, ECCV and NeurIPS.
\end{IEEEbiography}

\begin{IEEEbiography}[{\includegraphics[width=1in,height=1.25in,clip,keepaspectratio]{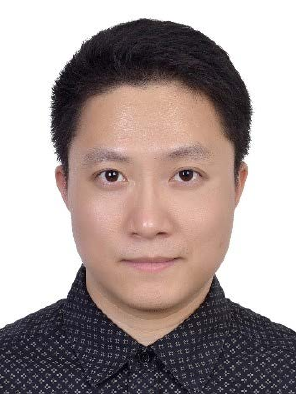}}]{Liang Lin}(M'09, SM'15) is a Full Professor of computer science at Sun Yat-Sen University. He served as the Executive Director and Distinguished Scientist of SenseTime Group from 2016 to 2018, leading the R\&D teams for cutting-edge technology transferring. He has authored or co-authored more than 200 papers in leading academic journals and conferences (e.g., 20+ papers in TPAMI/IJCV), and his papers have been cited by more than 16,000 times. He is an associate editor of IEEE Trans. Neural Networks and Learning Systems and IEEE Trans. Human-Machine Systems, and served as Area Chairs for numerous conferences such as CVPR, ICCV, SIGKDD and AAAI. He is the recipient of numerous awards and honors including Wu Wen-Jun Artificial Intelligence Award, the First Prize of China Society of Image and Graphics, ICCV Best Paper Nomination in 2019, Annual Best Paper Award by Pattern Recognition (Elsevier) in 2018, Best Paper Dimond Award in IEEE ICME 2017, Google Faculty Award in 2012. His supervised PhD students received ACM China Doctoral Dissertation Award, CCF Best Doctoral Dissertation and CAAI Best Doctoral Dissertation. He is a Fellow of IET.
\end{IEEEbiography}

\end{document}